\documentclass[journal]{IEEEtran}
\IEEEoverridecommandlockouts
\usepackage{cite}
\usepackage{amsmath,amssymb,amsfonts}
\usepackage{graphicx}
\usepackage{textcomp}
\usepackage{xcolor}
\usepackage{multirow}
\usepackage{algorithm}
\usepackage{stfloats}
\usepackage{algorithmic}
\usepackage{subcaption}
\usepackage{bm}
\usepackage[a4paper, total={184mm,239mm}]{geometry}

\def\BibTeX{{\rm B\kern-.05em{\sc i\kern-.025em b}\kern-.08em
    T\kern-.1667em\lower.7ex\hbox{E}\kern-.125emX}}

\begin{document}

\title{Neural Network-Assisted CLEAN for Efficient Multipath Parameter Estimation in MIMO Systems}

\author{
\IEEEauthorblockN{Chaofan Deng$^{1}$, Linyu Sun$^{1}$, Jaeho Lee$^{1}$, Zishen Wan$^{2}$, Arijit Raychowdhury$^{1}$}
\IEEEauthorblockA{\\$^{1}$Georgia Institute of Technology, Atlanta, GA, USA \hspace{2pt}
$^{2}$Columbia University, New York, NY, USA\\
\textit{Email: \{cdeng63, lsun327, jlee3941\}@gatech.edu, zishen.wan@columbia.edu, arijit.raychowdhury@ece.gatech.edu}}
}

\maketitle

\begin{abstract}
High-resolution multipath estimation in wideband multiple-input multiple-output (MIMO) channels requires repeated searches over angle of arrival, angle of departure, and propagation distance. This paper presents neural network-assisted CLEAN (NN-CLEAN), which replaces the multidimensional search at each extraction step with a multi-head classifier while retaining least-squares (LS) estimation of the complex path gain, parametric reconstruction, and residual cancellation. The same classifier is applied to successive residuals, allowing the extraction depth to vary without changing its output structure. With the number of iterations prescribed by the reference multipath-component (MPC) count, off-grid synthetic tests show that NN-CLEAN slightly exceeds grid-search CLEAN (GS-CLEAN) in path accuracy for one and two MPCs, although its accuracy declines more rapidly at larger path counts. For complete three-MPC extraction at a batch size of eight, NN-CLEAN achieves an approximately $27\times$ speedup and a $52$-fold reduction in peak GPU memory relative to GS-CLEAN. Applied without retraining to two-MPC line-of-sight (LOS) and non-line-of-sight (NLOS) Sionna ray-tracing channels within a coded MIMO link, NN-CLEAN substantially reduces channel-reconstruction mean-squared error (MSE) and bit error rate (BER) relative to interpolated LS channel estimation. GS-CLEAN retains an accuracy advantage on these channels, indicating the estimation cost associated with the computational savings.
\end{abstract}


\section{Introduction}

Accurate multipath channel characterization is essential to wideband
multiple-input multiple-output (MIMO) communication, channel sounding,
localization, and integrated sensing and communication (ISAC). Measurements
across multiple antennas and frequencies enable individual multipath
components (MPCs) to be distinguished through their spatial and
frequency-dependent signatures. Estimating physically meaningful parameters,
including the angle of arrival (AoA), angle of departure (AoD), propagation
distance or delay, and complex path gain, is therefore important for channel
reconstruction, beamforming, spatial multiplexing, and sensing. Reliable
parameter extraction remains difficult in complex propagation environments,
particularly at low signal-to-noise ratio (SNR), where noise and inter-path
interference obscure weak components and reduce the separability of closely
spaced paths \cite{ISAC_ROBUST,ISCAS_low_snr}.

Conventional high-resolution parameter estimation is predominantly
model based. Subspace methods such as MUSIC \cite{music,3d_music} and ESPRIT
\cite{esprit,3d_esprit} exploit the structure of the signal and noise
subspaces to estimate propagation parameters without an exhaustive
likelihood search. Their performance, however, depends on reliable
model-order selection and subspace separation. Correlated or coherent MPCs
can reduce the effective signal-subspace rank, while low-SNR observations
can cause subspace leakage or subspace swaps, resulting in substantial
estimation degradation \cite{hussain2023low,xu2022high}.

Maximum-likelihood estimation (MLE) provides an alternative model-based
formulation by identifying the propagation parameters that best explain the
observed channel under an assumed physical model. Although MLE does not rely
on the same subspace-separation conditions, joint estimation becomes
computationally prohibitive when several parameters and MPCs must be
optimized simultaneously. CLEAN \cite{seun_clean} reduces the joint problem
to a sequence of dominant-path estimation and subtraction steps. Each
iteration, however, generally requires a multidimensional parameter search,
causing the runtime and memory requirements to grow with the grid resolution,
number of search dimensions, and channel-observation size
\cite{sun2026guided,mle_low_snr}. This scaling limits the applicability of
grid-search CLEAN in systems requiring high-throughput or low-latency channel
processing.

Deep learning (DL) has been investigated as a means of reducing the
computational cost of multipath parameter estimation
\cite{NN1,NN2,NN3,NN4,NN5,NN_off_grid,NN_vary_MPC}. Existing approaches can
be broadly divided into direct DL estimators and DL-assisted model-based
methods. A direct DL estimator uses a neural network (NN) to map the received
signal or channel observation to the desired propagation parameters in one
forward pass. Although inference can be performed efficiently, simultaneous
estimation of a complete multipath channel remains difficult. A fixed output
structure must identify the active paths, preserve the association among the
parameters belonging to each path, and remain valid when the path count
differs from the training condition \cite{NN_vary_MPC}. Classification-based
estimators are further limited by their finite output grids and can degrade
when the true parameters lie between the predefined grid points
\cite{NN_off_grid}.

DL-assisted model-based methods instead replace selected computational
operations within an established signal-processing algorithm. This approach
retains the physical channel model and structured parameter updates while
using learned inference to reduce the cost of the most demanding operation.
Such a formulation is well suited to CLEAN, whose reconstruction and
successive-cancellation steps are analytically defined but whose repeated
multidimensional searches are computationally intensive. To the best of our
knowledge, neural inference has not previously been integrated directly into
the iterative CLEAN procedure for joint AoA, AoD, and propagation-distance
estimation.

This work proposes NN-assisted CLEAN (NN-CLEAN), a hybrid
multipath parameter estimator in which a multi-head classifier replaces the
exhaustive parameter search at each CLEAN iteration. The remaining operations
are model based: the complex path gain is obtained through least-squares (LS)
projection, the estimated MPC is reconstructed from the parametric channel
model, and the reconstructed response is subtracted from the current residual.
The updated residual is then supplied to the same single-MPC classifier for
the next extraction. Reusing one classifier across iterations permits different extraction
depths without changing the NN output dimension.

The replacement of grid search with NN inference also changes the
computational scaling under concurrent processing. Grid-search CLEAN must
evaluate the complete candidate space for every channel realization, whereas
NN-CLEAN can process multiple residual channels using batched tensor
operations on parallel hardware. The computational benefit is therefore
expected to become most significant when several channel realizations are
processed simultaneously, as may occur in centralized MIMO and ISAC systems
serving multiple communication links or sensing targets within the same
processing interval.

NN-CLEAN is evaluated using both statistically synthesized channels and
geometry-dependent channels generated with the NVIDIA Sionna ray-tracing
(RT) package. The synthetic experiments examine off-grid estimation over
different path counts, SNRs, antenna dimensions, and numbers of frequency
samples. Runtime and peak memory are measured as the concurrent processing
batch increases. The RT channels are incorporated into a coded MIMO
communication link, where path accuracy, channel-reconstruction mean-squared
error (MSE), and bit-error rate (BER) quantify how parameter-estimation errors
propagate to downstream equalization and decoding.

A preliminary conference paper introduced the core NN-CLEAN concept
under a controlled synthetic configuration \cite{deng2026neural}.
This journal article extends the evaluation to off-grid channels
with different path counts and SNRs, adds conditional parameter-precision
analysis, and examines observation dimensions and search-grid
resolution. GPU profiling distinguishes single-path search from
complete multipath extraction and measures runtime and peak memory
under increasing batch workloads. Two-MPC LOS and NLOS Sionna RT
channels further assess reconstruction and BER within a coded MIMO
link, while three-MPC LOS ablations examine label smoothing, angular
conditioning, distance-loss weighting, and the training curriculum.

The main contributions of this work are summarized as follows:
\begin{enumerate}

    \item \textbf{NN-Assisted CLEAN Framework:}
    We develop an NN-CLEAN estimator that replaces the exhaustive
    AoA--AoD--distance search within each CLEAN iteration with a multi-head
    neural classifier. The proposed method retains LS gain estimation,
    parametric channel reconstruction, and successive residual cancellation,
    providing a structured alternative to both exhaustive grid search and
    fully data-driven one-shot estimation.

    \item \textbf{Sequential Extraction with Variable Depth:}
    The same single-MPC classifier is applied repeatedly to the evolving
    channel residual. This formulation supports a variable extraction depth
    without changing the network output dimension and reduces the
    cross-path association burden encountered by one-shot multipath
    estimators. Its robustness is evaluated on off-grid synthetic channels across different path counts and SNRs,

    \item \textbf{Computational and System-Level Validation:}
    The proposed method is evaluated in terms of path accuracy, conditional
    parameter precision, channel-reconstruction fidelity, runtime, and peak
    memory. Geometry-dependent Sionna RT channels are further incorporated
    into a coded MIMO link, where channel-reconstruction MSE and BER are used
    to quantify the effect of parameter-estimation errors on downstream
    communication performance.

\end{enumerate}

\section{System Model and CLEAN Algorithm}
\label{sec:system_model}

\subsection{Multipath Channel Model}

\begin{figure}[t]
  \centering
  \includegraphics[width=\columnwidth]{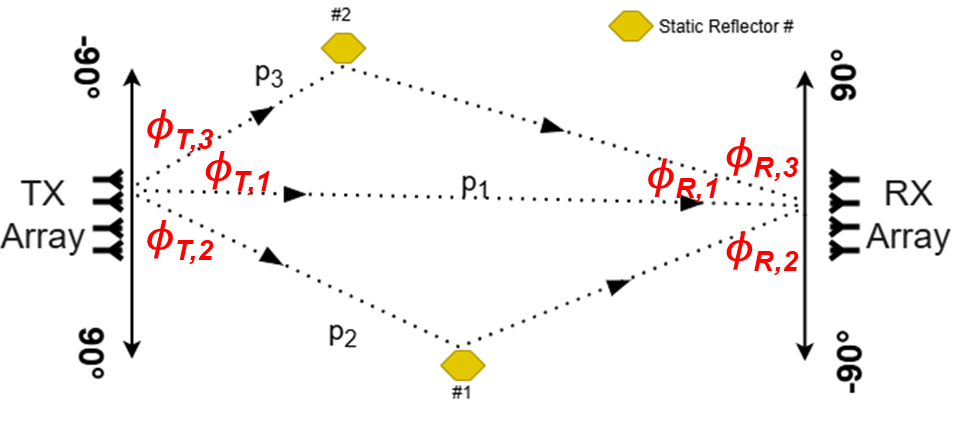}
  \caption{Channel Model Visualization}
  \label{fig:system_model}
\end{figure}

Consider a quasi-static double-directional radio channel between an
$N_T$-element transmitter (Tx) uniform linear array (ULA) and an
$N_R$-element receiver (Rx) ULA, as illustrated in
Fig.~\ref{fig:system_model}. The channel comprises
$N_{\mathrm{MPC}}$ resolvable multipath components (MPCs), including a
possible line-of-sight (LOS) component and reflected or scattered paths.
Each MPC is characterized by its angle of arrival (AoA), angle of departure
(AoD), propagation distance, and complex path gain.

The geometric parameters of an MPC are collected in

\begin{equation}
\label{eq:theta_definition}
\boldsymbol{\theta}
=
[\phi_R,\phi_T,d]^{\mathsf{T}},
\end{equation}
where $\phi_R$, $\phi_T$, and $d$ denote the azimuth AoA, azimuth AoD, and
propagation distance, respectively.

Let $f_k$, $k=1,\ldots,N_f$, denote the $k$th frequency point. Following
the double-directional channel formulation in \cite{seun_clean}, let
$b_T(\phi_T,f_k)\in\mathbb{C}^{N_T\times1}$ and
$b_R(\phi_R,f_k)\in\mathbb{C}^{1\times N_R}$ denote the calibrated
vertically polarized beam-pattern responses of the Tx and Rx arrays,
respectively. The spatial--frequency response associated with a candidate
MPC is then defined as

\begin{equation}
\label{eq:path_response}
\boldsymbol{P}(\boldsymbol{\theta},f_k)
=
b_T(\phi_T,f_k)b_R(\phi_R,f_k)
\exp\!\left(-j2\pi f_k\frac{d}{c}\right),
\end{equation}
where $c$ is the speed of light and
$\boldsymbol{P}(\boldsymbol{\theta},f_k)
\in\mathbb{C}^{N_T\times N_R}$.

The channel transfer function between the Tx and Rx arrays at frequency
$f_k$ is modeled as the superposition of the individual MPC responses:

\begin{equation}
\label{eq:channel_model}
\boldsymbol{H}(f_k)
=
\sum_{l=1}^{N_{\mathrm{MPC}}}
\alpha_l
\boldsymbol{P}^{\mathsf{T}}(\boldsymbol{\theta}_l,f_k)
+
\boldsymbol{N}(f_k),
\qquad
1\leq k\leq N_f,
\end{equation}
where $\boldsymbol{\theta}_l$ and
$\alpha_l\in\mathbb{C}$ denote the geometric parameter vector and complex
path gain of the $l$th MPC, respectively. The transpose establishes the
receive-by-transmit channel orientation, such that
$\boldsymbol{H}(f_k)\in\mathbb{C}^{N_R\times N_T}$. The matrix
$\boldsymbol{N}(f_k)\in\mathbb{C}^{N_R\times N_T}$ represents AWGN with
independent and identically distributed complex Gaussian entries.

\subsection{MLE-Based MPC Extraction Using CLEAN}

Under the AWGN assumption, multipath parameter estimation can be formulated
as a deterministic maximum-likelihood estimation (MLE) problem. For a known
number of MPCs, maximizing the likelihood is equivalent to minimizing the
reconstruction error between the observed channel and the noise-free
parametric component of \eqref{eq:channel_model}. The joint estimator is
therefore given by

\begin{equation}
\label{eq:joint_ml}
\begin{aligned}
\{
\hat{\alpha}_l,\hat{\boldsymbol{\theta}}_l
\}_{l=1}^{N_{\mathrm{MPC}}}
&=
\operatorname*{arg\,min}_{\{\alpha_l,\boldsymbol{\theta}_l\}}
\\[-1mm]
&\quad
\sum_{k=1}^{N_f}
\left\|
\boldsymbol{H}(f_k)
-
\sum_{l=1}^{N_{\mathrm{MPC}}}
\alpha_l
\boldsymbol{P}^{\mathsf{T}}(\boldsymbol{\theta}_l,f_k)
\right\|_F^2 ,
\end{aligned}
\end{equation}
where $\|\cdot\|_F$ denotes the Frobenius norm. Directly solving
\eqref{eq:joint_ml} requires simultaneous optimization over the geometric
parameters and complex gains of all MPCs, resulting in a high-dimensional
nonlinear problem whose complexity increases rapidly with the path count.

CLEAN provides a practical alternative by decomposing the joint problem
into a sequence of single-MPC estimation steps
\cite{seun_clean,clean_algorithm}. At each iteration, the dominant
component in the current residual is estimated, reconstructed, and
subtracted before the next component is extracted. In this work, the CLEAN
implementation that identifies each dominant MPC through an exhaustive
discretized search over AoA, AoD, and distance is referred to as
grid-search CLEAN (GS-CLEAN).

Let $\boldsymbol{H}_{\mathrm{rem}}^{(i)}(f_k)$ denote the residual channel
at extraction iteration $i$. The residual is initialized using the observed
channel as

\begin{equation}
\label{eq:residual_init}
\boldsymbol{H}_{\mathrm{rem}}^{(1)}(f_k)
=
\boldsymbol{H}(f_k).
\end{equation}
In GS-CLEAN, the AoA, AoD, and distance dimensions are discretized over
$\Theta_{\mathrm{AoA}}$, $\Theta_{\mathrm{AoD}}$, and
$\Theta_{\mathrm{dist}}$, containing $N_{\phi_R}$, $N_{\phi_T}$, and
$N_d$ candidate values, respectively. The resulting joint search space is

\begin{equation}
\label{eq:search_space}
\begin{aligned}
\Theta
&=
\Theta_{\mathrm{AoA}}
\times
\Theta_{\mathrm{AoD}}
\times
\Theta_{\mathrm{dist}},
\\[-1mm]
|\Theta|
&=
N_{\phi_R}N_{\phi_T}N_d .
\end{aligned}
\end{equation}

Under the normalized array responses adopted in this work, the energy of
the parametric response is constant across the search grid. Consequently,
maximizing the concentrated single-path likelihood reduces to evaluating
the spatial--frequency correlation between each candidate response and the
current residual. The dominant MPC parameter vector at iteration $i$ is
estimated as

\begin{equation}
\label{eq:ml_estimate}
\hat{\boldsymbol{\theta}}_i
=
\operatorname*{arg\,max}_{\boldsymbol{\theta}\in\Theta}
\left|
\sum_{k=1}^{N_f}
\left\langle
\boldsymbol{P}^{\mathsf{T}}(\boldsymbol{\theta},f_k),
\boldsymbol{H}_{\mathrm{rem}}^{(i)}(f_k)
\right\rangle_F
\right|^2 ,
\end{equation}
where
$\langle\boldsymbol{A},\boldsymbol{B}\rangle_F
=\operatorname{tr}(\boldsymbol{A}^{H}\boldsymbol{B})$
denotes the Frobenius inner product. After estimating
$\hat{\boldsymbol{\theta}}_i$, the corresponding complex path gain is
obtained through least-squares (LS) projection:

\begin{equation}
\label{eq:path_gain}
\hat{\alpha}_i
=
\frac{
\displaystyle
\sum_{k=1}^{N_f}
\left\langle
\boldsymbol{P}^{\mathsf{T}}
(\hat{\boldsymbol{\theta}}_i,f_k),
\boldsymbol{H}_{\mathrm{rem}}^{(i)}(f_k)
\right\rangle_F
}{
\displaystyle
\sum_{k=1}^{N_f}
\left\|
\boldsymbol{P}^{\mathsf{T}}
(\hat{\boldsymbol{\theta}}_i,f_k)
\right\|_F^2
}.
\end{equation}
The estimated MPC is then reconstructed and subtracted from the current
residual according to

\begin{equation}
\label{eq:residual}
\boldsymbol{H}_{\mathrm{rem}}^{(i+1)}(f_k)
=
\boldsymbol{H}_{\mathrm{rem}}^{(i)}(f_k)
-
\hat{\alpha}_i
\boldsymbol{P}^{\mathsf{T}}
(\hat{\boldsymbol{\theta}}_i,f_k).
\end{equation}

The parameter search, gain estimation, and residual subtraction in
\eqref{eq:ml_estimate}--\eqref{eq:residual} are repeated for subsequent
MPCs. Extraction may terminate after a prescribed number of components or
when an external stopping criterion is satisfied, such as when the
estimated path gain or residual energy falls below a predefined threshold.

\subsection{Computational Bottleneck of GS-CLEAN}

The computational cost of GS-CLEAN is dominated by the exhaustive parameter
evaluation in \eqref{eq:ml_estimate}. For a single candidate
$\boldsymbol{\theta}$, evaluating the spatial--frequency correlation over
$N_f$ observations of an $N_R\times N_T$ channel requires approximately

\begin{equation}
\mathcal{O}(N_fN_RN_T)
\end{equation}
operations. Because the AoA--AoD--distance search space contains
$N_{\phi_R}N_{\phi_T}N_d$ candidate combinations, the resulting
per-iteration complexity scales as

\begin{equation}
\label{eq:gsclean_complexity}
\mathcal{O}\!\left(
N_{\phi_R}N_{\phi_T}N_d
N_fN_RN_T
\right).
\end{equation}

For comparable grid sizes of approximately $G$ points per parameter
dimension, this dependence reduces to
$\mathcal{O}(G^3N_fN_RN_T)$. Because the exhaustive search is repeated
after each MPC subtraction, the total computational burden also increases
approximately linearly with the number of extracted components for a fixed
search space and observation dimension.

Repeated computation can be reduced by precomputing candidate responses or
array-manifold dictionaries. This strategy, however, introduces a memory
requirement that grows with both the number of candidate combinations and
the spatial--frequency observation dimensions. Increasing the grid
resolution, antenna count, or number of frequency samples therefore
increases the computational workload, the required dictionary storage, or
both.

The limitation becomes more pronounced when additional propagation
parameters are incorporated. The present formulation considers a
three-dimensional search over azimuth AoA, azimuth AoD, and distance.
Including Tx and Rx elevation angles would expand the search space to five
dimensions, resulting in approximately
$\mathcal{O}(G^5N_fN_RN_T)$ complexity for comparable grid resolutions.
Introducing Doppler for a time-varying channel adds another search
dimension and further increases the number of candidate hypotheses.
Consequently, exhaustive GS-CLEAN becomes progressively more
computationally and memory intensive as the parameter space and observation
dimensions increase.

\section{Proposed NN-CLEAN Algorithm}
\label{sec:nn_clean}

\subsection{NN-Assisted Iterative MPC Extraction}

\begin{figure}[htbp]
  \centering
  \includegraphics[width=\columnwidth]{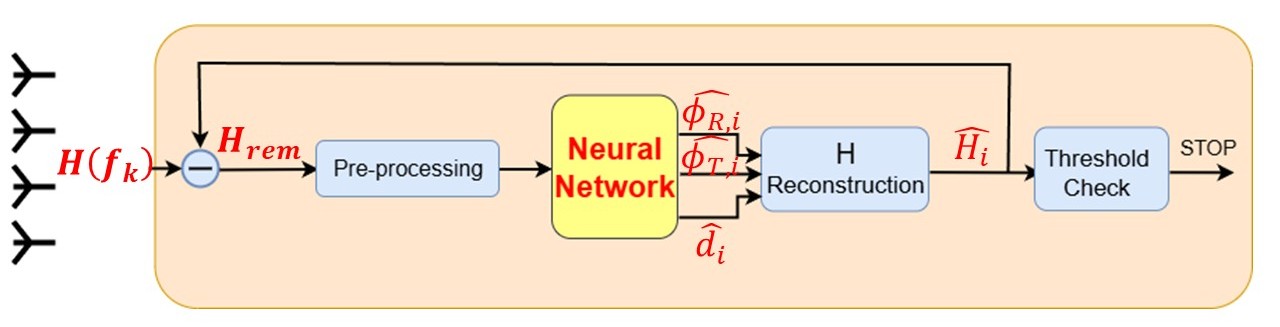}
  \caption{Overview of the proposed NN-CLEAN framework.}
  \label{fig:2}
\end{figure}

To address the multidimensional search bottleneck identified in
Section~\ref{sec:system_model}, we propose NN-CLEAN, which integrates neural
network (NN) inference into the iterative CLEAN framework. As illustrated
in Fig.~\ref{fig:2}, the exhaustive parameter search in
\eqref{eq:ml_estimate} is replaced by a direct mapping from the current
residual channel $\boldsymbol{H}_{\mathrm{rem}}^{(i)}(f_k)$ to the
geometric parameters of the dominant remaining MPC. The model-based path-gain
estimation, parametric reconstruction, and residual-subtraction operations
of conventional CLEAN are retained.

At the $i$th iteration, the residual channel observations are supplied
directly to the NN, which produces the dominant MPC parameter estimate
$\hat{\boldsymbol{\theta}}_i$. Given this estimate, the corresponding
complex path gain $\hat{\alpha}_i$ is computed using the LS projection in
\eqref{eq:path_gain}. The estimated MPC is then reconstructed and removed
from the residual according to \eqref{eq:residual}, after which
$\boldsymbol{H}_{\mathrm{rem}}^{(i+1)}(f_k)$ is supplied to the same
network for the next extraction iteration. NN-CLEAN therefore replaces only
the computationally intensive grid-search operation while preserving the
physical reconstruction and successive-cancellation structure of CLEAN.

An iterative single-MPC architecture is adopted instead of a one-shot
network that estimates all MPCs simultaneously. A one-shot formulation must
maintain the association among the AoA, AoD, and distance estimates
belonging to each MPC, and this association becomes increasingly difficult
as the path count increases. Moreover, a one-shot estimator generally
requires a predefined output structure for the complete multipath set,
making it less suitable for variable path counts. In contrast, NN-CLEAN
repeatedly applies the same single-MPC estimator to the evolving residual,
allowing the extraction depth to be prescribed or controlled by an external
CLEAN stopping criterion without changing the NN output dimension.

\subsection{Proposed NN Architecture}

\begin{figure}[htbp]
  \centering
  \includegraphics[width=\columnwidth]{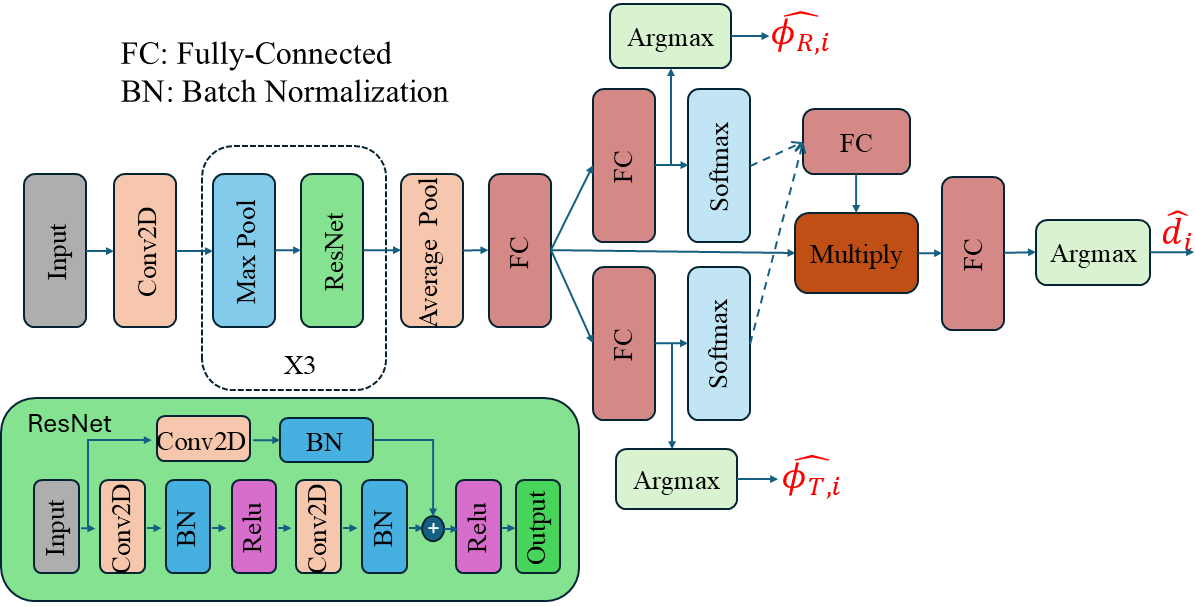}
  \caption{Multi-head residual classification architecture used in
  NN-CLEAN.}
  \label{fig:3}
\end{figure}

The NN architecture used for MPC parameter estimation is illustrated in
Fig.~\ref{fig:3}. A shared feature-extraction backbone processes the
residual channel and learns a common representation of its
spatial--frequency characteristics. The extracted representation is then
provided to parameter-specific classification heads for AoA, AoD, and
distance estimation.

The shared backbone comprises pooling and residual blocks that progressively
extract the relevant channel features while reducing the intermediate
feature dimension. The resulting representation is supplied to the AoA and
AoD classification branches. The distance branch additionally employs a
fully connected (FC) gating mechanism conditioned on the angular outputs,
allowing information from the estimated angular dimensions to contribute to
distance classification.
Rather than treating every element of the Cartesian search space $\Theta$
as a separate class, the network produces three unnormalized class-logit
vectors at extraction iteration $i$:

\begin{equation}
\label{eq:nn_outputs}
\hat{\boldsymbol{y}}_{\phi_R}^{(i)}
\in\mathbb{R}^{N_{\phi_R}},
\qquad
\hat{\boldsymbol{y}}_{\phi_T}^{(i)}
\in\mathbb{R}^{N_{\phi_T}},
\qquad
\hat{\boldsymbol{y}}_{d}^{(i)}
\in\mathbb{R}^{N_d}.
\end{equation}

Each logit vector assigns a classification score to the candidate grid
points of its corresponding parameter dimension. The estimated class index
for each output head is selected directly from the logits according to

\begin{equation}
\label{eq:nn_decoding}
\begin{aligned}
\hat n_{q,i}
&=
\operatorname*{arg\,max}_{n}
\hat y_{q,n}^{(i)},
\qquad
q\in\{\phi_R,\phi_T,d\},
\\
\hat{\boldsymbol{\theta}}_i
&=
\begin{bmatrix}
\Theta_{\mathrm{AoA}}[\hat n_{\phi_R,i}]\\
\Theta_{\mathrm{AoD}}[\hat n_{\phi_T,i}]\\
\Theta_{\mathrm{dist}}[\hat n_{d,i}]
\end{bmatrix}.
\end{aligned}
\end{equation}

The resulting estimate $\hat{\boldsymbol{\theta}}_i$ follows the same
parameterization as the estimate produced by GS-CLEAN and can therefore be
used directly in the path-gain estimation and residual-subtraction
operations in \eqref{eq:path_gain} and \eqref{eq:residual}. The multi-head
formulation also avoids constructing a single classifier over all
$N_{\phi_R}N_{\phi_T}N_d$ joint grid combinations. Instead, the network
produces a total of $N_{\phi_R}+N_{\phi_T}+N_d$ output logits.

\subsection{Computational Characteristics}

The principal computational distinction between GS-CLEAN and NN-CLEAN lies
in the estimation of $\hat{\boldsymbol{\theta}}_i$. At every extraction
iteration, GS-CLEAN evaluates
$N_{\phi_R}N_{\phi_T}N_d$ joint parameter combinations, whereas NN-CLEAN
obtains the three parameter estimates through a single forward pass with
factorized classification outputs. NN-CLEAN consequently avoids explicit
evaluation and storage of the complete multidimensional candidate space.

NN inference primarily comprises tensor operations that can be efficiently
parallelized across multiple channel realizations on modern accelerators.
The resulting runtime and memory characteristics are evaluated
experimentally in Section~\ref{sec:results} across different processing
batch sizes and system configurations.

\section{Simulation and Evaluation Methodology}
The evaluation uses the known number of MPCs retained in the reference
channel, $N_{\mathrm{MPC}}$, to prescribe the extraction depth of both
NN-CLEAN and GS-CLEAN. Only the path count is supplied; the geometric
parameters and complex gains are estimated from the channel observation.
Unequal reference and estimated path counts would require additional
association rules for unmatched components, while reconstruction MSE and
BER would combine model-order and parameter-estimation errors. Using a
common extraction depth therefore focuses the comparison on parameter
estimation and successive cancellation, with incorrect path selections
and their resulting reconstruction errors still included.

When the path count is unknown, the extraction depth must be determined
from residuals containing physical MPCs, observation noise, and
cancellation artifacts. The appropriate stopping criterion depends on
SNR, relative path powers, and estimator-dependent residual statistics,
making a fixed threshold difficult to apply across operating conditions.
Adaptive stopping criteria and their effects on missed and spurious
detections, reconstruction quality, BER, and computational cost are left
for future work.

\subsection{Synthetic MIMO Configuration}

The proposed NN-CLEAN framework is first evaluated using synthesized
wideband MIMO channels following the parametric model in
Section~\ref{sec:system_model}. Unless otherwise specified, a
$4\times4$ MIMO configuration with $N_T=N_R=4$ is considered over a
$1$~GHz bandwidth centered at $f_c=2.5$~GHz. The channel is sampled at
$N_f=512$ uniformly spaced frequency points. Both the Tx and Rx employ ULAs
with half-wavelength inter-element spacing evaluated at $f_c$ and ideal
vertically polarized omnidirectional elements with negligible mutual
coupling. The common system and search-space parameters are summarized in
Table~\ref{tab:system_configuration}.

For an $N$-element ULA, the frequency-dependent steering vector at azimuth
angle $\phi$ and frequency $f_k$ is

\begin{equation}
\label{eq:steering_vector}
\boldsymbol{b}(\phi,f_k)
=
\frac{1}{\sqrt{N}}
\left[
1,\,
e^{-j\pi\frac{f_k}{f_c}\sin(\phi)},\,
\ldots,\,
e^{-j\pi(N-1)\frac{f_k}{f_c}\sin(\phi)}
\right]^{\mathsf{T}}.
\end{equation}

For grid-based estimation and NN classification, the parameter space
$\Theta$ is discretized as follows:
\begin{itemize}
    \item \textbf{AoD ($\phi_T$) and AoA ($\phi_R$):}
    $[-90^\circ,90^\circ]$ with $1^\circ$ spacing, giving
    $N_{\phi_R}=N_{\phi_T}=181$.
    \item \textbf{Distance ($d$):}
    $[5,15]$~m with $0.05$~m spacing, giving $N_d=201$.
\end{itemize}

The complex path gains are generated independently according to
$\alpha_l\sim\mathcal{CN}(0,2)$. Let
$\boldsymbol{H}_0(f_k)$ denote the noise-free component of the synthetic
channel in \eqref{eq:channel_model}, such that
$\boldsymbol{H}(f_k)=\boldsymbol{H}_0(f_k)+\boldsymbol{N}(f_k)$. The
synthetic channel-observation SNR is defined as

\begin{equation}
\label{eq:synthetic_snr}
\mathrm{SNR}_{\mathrm{syn}}
=
10\log_{10}\!\left(
\frac{
\displaystyle\sum_{k=1}^{N_f}
\left\|\boldsymbol{H}_0(f_k)\right\|_F^2
}{
\displaystyle\sum_{k=1}^{N_f}
\left\|\boldsymbol{N}(f_k)\right\|_F^2
}
\right).
\end{equation}

Training and testing use different parameter distributions. During
training, the AoA, AoD, and distance values are restricted to the predefined
grids, corresponding to an \textbf{on-grid} configuration. During synthetic
evaluation, these parameters are sampled continuously over the same physical
ranges, resulting in an \textbf{off-grid} test condition. Consequently,
both NN-CLEAN and GS-CLEAN are evaluated in the presence of grid mismatch
rather than only at parameter values represented exactly by the search
dictionary.
\begin{table}[t]
\centering
\caption{Common Simulation Parameters}
\label{tab:system_configuration}
\footnotesize
\renewcommand{\arraystretch}{1.10}
\begin{tabular}{p{0.42\columnwidth} p{0.48\columnwidth}}
\hline
\textbf{Parameter} & \textbf{Setting} \\
\hline
Center frequency, $f_c$
    & $2.5$~GHz \\
Bandwidth, $B$
    & $1$~GHz \\
Subcarriers, $N_f$
    & $512$ \\
Tx/Rx elements, $N_T,N_R$
    & $4,4$ \\
AoA/AoD, $\phi_R,\phi_T$
    & $[-90^\circ,90^\circ]$, $1^\circ$ resolution \\
Distance, $d$
    & $[5,15]$~m, $0.05$~m resolution \\
\hline
\end{tabular}
\end{table}

\subsection{NN Training Procedure}
\label{sec:training_setup}

Training samples are synthesized on the fly during optimization. For each
sample, the path count is selected from
$N_{\mathrm{MPC}}\in\{1,2,3,4\}$, and the corresponding AoA, AoD, and
distance parameters are drawn from their respective on-grid sets. The
synthetic channel-observation SNR is independently sampled from
$[-5,10]$~dB. The network parameters $\boldsymbol{w}$ are optimized using
Adam with a learning rate of $10^{-3}$ and a batch size of $128$.

Because NN-CLEAN extracts MPCs sequentially, the ground-truth paths are
ordered in descending magnitude of their complex gains. The target at
extraction iteration $i$ therefore corresponds to the $i$th strongest MPC.
Gaussian label smoothing (GLS) \cite{GLS} is applied to account for the
correlation among neighboring parameter bins. For parameter head
$q\in\{\phi_R,\phi_T,d\}$, the GLS target at class index $n$ is

\begin{equation}
\label{eq:gls}
t_{q,n}^{(i)}
=
\frac{
\exp\!\left[
-\dfrac{1}{2}
\left(
\dfrac{n-\mu_q^{(i)}}{\sigma}
\right)^2
\right]
}{
\displaystyle
\sum_{m=1}^{N_q}
\exp\!\left[
-\dfrac{1}{2}
\left(
\dfrac{m-\mu_q^{(i)}}{\sigma}
\right)^2
\right]
},
\end{equation}
where $\mu_q^{(i)}$ is the ground-truth class index of the $i$th extracted
MPC, $N_q$ is the number of classes associated with head $q$, and
$\sigma=2.0$ is the smoothing parameter.

\begin{algorithm}[htbp]
\caption{Three-Phase Training Procedure for NN-CLEAN}
\label{alg:model-training}
\begin{algorithmic}[1]
\STATE \textbf{Initialize:} Model weights $\boldsymbol{w}$ and Adam optimizer
       with $\mathrm{lr}=10^{-3}$
\FOR{each training step $t=1$ \textbf{to} $1{,}000{,}000$}
    \STATE Sample
    $\mathrm{SNR}_{\mathrm{syn}}\sim\mathcal{U}(-5,10)$~dB
    \STATE Sample $N_{\mathrm{MPC}}\in\{1,2,3,4\}$
    \STATE Generate
    $\{\phi_{R,l},\phi_{T,l},d_l\}_{l=1}^{N_{\mathrm{MPC}}}\in\Theta$
    \STATE Sample $\alpha_l\sim\mathcal{CN}(0,2)$
    \STATE Synthesize $\boldsymbol{H}_0$ and add AWGN to obtain
           $\boldsymbol{H}$
    \STATE Sort the MPCs in descending order of $|\alpha_l|$
    \STATE Generate the GLS targets using \eqref{eq:gls}

    \IF{$t\leq500{,}000$}
        \STATE \textit{// Phase 1: Dominant-path estimation}
        \STATE Evaluate $\mathcal{L}_{\mathrm{final}}$ using the
               Phase~1 case of \eqref{eq:phase_loss}

    \ELSIF{$t\leq750{,}000$}
        \STATE \textit{// Phase 2: Teacher-forced extraction}
        \STATE Extract $N_{\mathrm{MPC}}$ paths using ground-truth
               residual subtraction
        \STATE Evaluate $\mathcal{L}_{\mathrm{final}}$ using the
               Phase~2 case of \eqref{eq:phase_loss}

    \ELSE
        \STATE \textit{// Phase 3: Autoregressive extraction}
        \STATE Extract $N_{\mathrm{MPC}}$ paths using predicted parameters
               and LS gain estimates
        \STATE Evaluate $\mathcal{L}_{\mathrm{final}}$ using the
               Phase~3 case of \eqref{eq:phase_loss}
    \ENDIF

    \STATE Backpropagate $\mathcal{L}_{\mathrm{final}}$ and update
           $\boldsymbol{w}$
\ENDFOR
\end{algorithmic}
\end{algorithm}

At each extraction iteration, the network operates on a different residual
channel. A separate classification loss is therefore evaluated for each
residual state encountered during sequential extraction. Let
$\boldsymbol{t}_q^{(i)}$ denote the GLS target vector and
$\hat{\boldsymbol{y}}_q^{(i)}$ denote the corresponding unnormalized logit
vector. The categorical cross-entropy (CCE) for parameter head $q$ is

\begin{equation}
\label{eq:cce_definition}
\begin{aligned}
\mathrm{CCE}\!\left(
\boldsymbol{t}_q^{(i)},
\hat{\boldsymbol{y}}_q^{(i)}
\right)
=
-\sum_{n=1}^{N_q}
t_{q,n}^{(i)}
\log\!\left(
\frac{
\exp\!\left(\hat y_{q,n}^{(i)}\right)
}{
\displaystyle
\sum_{m=1}^{N_q}
\exp\!\left(\hat y_{q,m}^{(i)}\right)
}
\right).
\end{aligned}
\end{equation}

The loss associated with extraction iteration $i$ is defined as

\begin{equation}
\label{eq:step_loss}
\begin{aligned}
\mathcal{L}_{\mathrm{iter}}^{(i)}
={}&
\mathrm{CCE}\!\left(
\boldsymbol{t}_{\phi_R}^{(i)},
\hat{\boldsymbol{y}}_{\phi_R}^{(i)}
\right)
+
\mathrm{CCE}\!\left(
\boldsymbol{t}_{\phi_T}^{(i)},
\hat{\boldsymbol{y}}_{\phi_T}^{(i)}
\right)
\\
&+
3\,
\mathrm{CCE}\!\left(
\boldsymbol{t}_{d}^{(i)},
\hat{\boldsymbol{y}}_{d}^{(i)}
\right).
\end{aligned}
\end{equation}

A weight of $3$ is applied exclusively to the distance loss because a
distance error produces a frequency-dependent phase mismatch in the
reconstructed MPC. The resulting coherent residual can propagate through
subsequent CLEAN iterations and impair the extraction of weaker components.

The final training loss depends on the curriculum phase:

\begin{equation}
\label{eq:phase_loss}
\mathcal{L}_{\mathrm{final}}
=
\begin{cases}
\mathcal{L}_{\mathrm{iter}}^{(1)},
& \text{Phase 1},\\[2mm]
\displaystyle
\frac{1}{N_{\mathrm{MPC}}}
\sum_{i=1}^{N_{\mathrm{MPC}}}
\mathcal{L}_{\mathrm{iter}}^{(i)},
& \text{Phases 2 and 3}.
\end{cases}
\end{equation}

Training comprises $10^6$ optimization steps divided into three phases.
Phase~1 uses the first $50\%$ steps to train dominant-path estimation.
Phase~2 uses the next $25\%$ steps to introduce sequential extraction
with ground-truth residual subtraction. Phase~3 uses the final $25\%$
steps to generate the residual sequence from the predicted parameters and
their LS gain estimates. Thus, Phases~2 and~3 use the same loss aggregation
in \eqref{eq:phase_loss}, but differ in how the residual channels are
formed. During Phase~3, the hard parameter decisions and residual updates
are treated as stop-gradient operations.

\subsection{Evaluation Protocol}
\label{sec:evaluation_protocol}

\subsubsection{Baselines and Estimation Metrics}

The primary synthetic evaluation is conducted entirely under off-grid
testing, with the ground-truth AoA, AoD, and distance drawn continuously
within the ranges in Table~\ref{tab:system_configuration}. NN-CLEAN is
compared with GS-CLEAN \cite{seun_clean}, 3D-MUSIC \cite{3d_music},
3D-ESPRIT \cite{3d_esprit}, and a standalone one-shot NN estimator.
GS-CLEAN serves as the exhaustive grid-search reference, while the subspace
methods are evaluated only for path counts supported by the available
signal-subspace dimension.

The one-shot baseline uses the same network architecture, parameter
grids, loss weights, and optimization settings as NN-CLEAN. A single
model is trained with varying MPC counts from
$N_{\mathrm{MPC}}\in\{1,2,3,4\}$. For each parameter head, the training
target is formed by summing Gaussian kernels centered at the
ground-truth bins of all MPCs and normalizing the resulting vector.
Training uses the original multipath observation throughout, without
the sequential residual updates or three-phase curriculum of
NN-CLEAN. After a single forward pass, the
$N_{\mathrm{MPC}}$ highest-scoring bins are selected independently
from each head using the known reference path count. Ideal association
among the selected AoA, AoD, and distance values is assumed during
evaluation.

For NN-CLEAN and GS-CLEAN, the reference MPCs are ranked in descending
order of their complex-gain magnitudes. The MPC estimated at extraction
iteration $i$ is compared with the $i$th reference MPC in this ordering.
Unless otherwise specified, a paired MPC is counted as correct only
when its AoA, AoD, and distance estimates all lie within $\pm2$
corresponding grid intervals of the ground truth. Accuracy is
defined as

\begin{equation}
\label{eq:path_accuracy}
A
=
\frac{N_{\mathrm{correct}}}{N_{\mathrm{true}}},
\end{equation}
where $N_{\mathrm{correct}}$ and $N_{\mathrm{true}}$ denote the numbers of
correctly estimated and ground-truth MPCs, respectively.

The precision of successfully associated MPCs is further characterized
using the angular and distance multipath-component distance (MCD). With the
angles expressed in radians, the angular MCD is

\begin{equation}
\label{eq:angular_mcd}
\eta_{\mathrm{ang}}
=
\frac{1}{2}
\sqrt{
\left|
e^{j\phi_R}-e^{j\hat{\phi}_R}
\right|^2
+
\left|
e^{j\phi_T}-e^{j\hat{\phi}_T}
\right|^2
},
\end{equation}
and the normalized distance MCD is

\begin{equation}
\label{eq:distance_mcd}
\eta_d
=
\frac{|d-\hat d|}{\Delta d_{\max}},
\qquad
\Delta d_{\max}=d_{\max}-d_{\min}.
\end{equation}

For either metric, the root-mean-square error (RMSE) over the $N_m$
successfully associated MPCs is

\begin{equation}
\label{eq:mcd_rmse}
\mathrm{RMSE}_{x}
=
\sqrt{
\frac{1}{N_m}
\sum_{l=1}^{N_m}
\eta_{x,l}^{2}
},
\qquad
x\in\{\mathrm{ang},d\}.
\end{equation}

The fidelity of the synthetic channel reconstruction is additionally
characterized by the energy-captured ratio

\begin{equation}
\label{eq:energy_ratio}
\epsilon
=
\frac{
\left\|
\displaystyle\sum_{i=1}^{N_{\mathrm{MPC}}}
\hat{\boldsymbol{H}}_i
\right\|_F^2
}{
\left\|\boldsymbol{H}_0\right\|_F^2
},
\end{equation}
where the Frobenius norms are evaluated over the complete
spatial--frequency channel. Using the noise-free reference
$\boldsymbol{H}_0$ prevents AWGN energy from being interpreted as
recoverable multipath energy.

\subsubsection{Runtime and Memory Profiling}
The computational comparison between NN-CLEAN and GS-CLEAN uses
synthesized channels under the baseline $4\times4$, $N_f=512$
configuration on an NVIDIA GeForce RTX~4070 GPU with 12~GB of device
memory. Runtime and peak GPU memory are measured as the processing
batch size increases from $1$ to $8$ for two workloads. The one-MPC
workload measures the parameter-search call from the observed channel
$\boldsymbol{H}$ to the estimator outputs, using NN inference for
NN-CLEAN and exhaustive AoA--AoD--distance search for GS-CLEAN. The
three-MPC workload measures the end-to-end extraction call, including
parameter search, LS gain estimation, path reconstruction, and residual
updates across three sequential iterations. Untimed warm-up runs are
performed before measurement to exclude graph tracing and one-time
initialization overhead.

Batch size specifies the number of independent channel realizations
processed concurrently within the same estimation call, and the reported
runtime covers the complete batch. Such a workload may arise in a
centralized MIMO or ISAC processor handling observations from multiple
users or vehicles within the same processing interval. Each sweep keeps
the path count fixed at one or three MPCs per realization, so increasing
batch size changes the concurrent workload while preserving the
extraction depth.

\subsubsection{Observation-Dimension and Search-Grid Resolution Sweeps}
\label{sec:parameter_sweeps}

The dependence of path accuracy on the observation dimensions and
search-grid resolution is evaluated using $N_{\mathrm{MPC}}=3$ off-grid
channels at $\mathrm{SNR}_{\mathrm{syn}}=5$~dB, with AoA, AoD, and
distance sampled continuously in each sweep. The antenna sweep varies
$N_T=N_R$ from $4$ to $8$ with $N_f=512$, and the subcarrier sweep
varies $N_f\in\{64,128,256,512,1024,2048\}$ with $N_T=N_R=4$; both
use the baseline search grid. The search-grid resolution sweep holds
$N_T=N_R=4$ and $N_f=512$ and uses the five grids in
Table~\ref{tab:search_grid_sweep}. The remaining system parameters follow
Table~\ref{tab:system_configuration}.

A separate NN-CLEAN model is trained for each configuration following
Algorithm~\ref{alg:model-training}. The antenna and subcarrier sweeps
change its input dimension, whereas the search-grid resolution sweep
changes only the sizes of its AoA, AoD, and distance classification
outputs and their directly connected layers. The remaining architecture
and training procedure are unchanged. For the corresponding GS-CLEAN
evaluations, the array and frequency responses are regenerated for each
observation dimension, and the search dictionary is constructed at each
grid resolution. Both methods are evaluated using the path accuracy
defined in \eqref{eq:path_accuracy}. The antenna and subcarrier sweeps use the
$\pm2$-grid tolerance of the baseline search grid. This same physical
tolerance, $\pm2^\circ$ for AoA and AoD and $\pm0.10$~m for distance,
is retained across all five resolution settings so that the permitted
parameter error does not vary with the search grid.

\begin{table}[t]
\centering
\caption{Search Grids Used in the Resolution Sweep}
\label{tab:search_grid_sweep}
\footnotesize
\renewcommand{\arraystretch}{1.10}
\begin{tabular}{lcc}
\hline
\textbf{Search grid} & \textbf{Angular resolution} &
\textbf{Distance resolution} \\
\hline
Very coarse & $4^\circ$ & $0.20$~m \\
Coarse & $2^\circ$ & $0.10$~m \\
Baseline & $1^\circ$ & $0.05$~m \\
Fine & $0.5^\circ$ & $0.025$~m \\
Very fine & $0.25^\circ$ & $0.0125$~m \\
\hline
\end{tabular}
\end{table}

\subsection{Sionna Ray-Tracing Communication-Link Configuration}
\label{sec:rt_setup}

\begin{figure*}[t!]
    \centering
    \includegraphics[width=\textwidth]
    {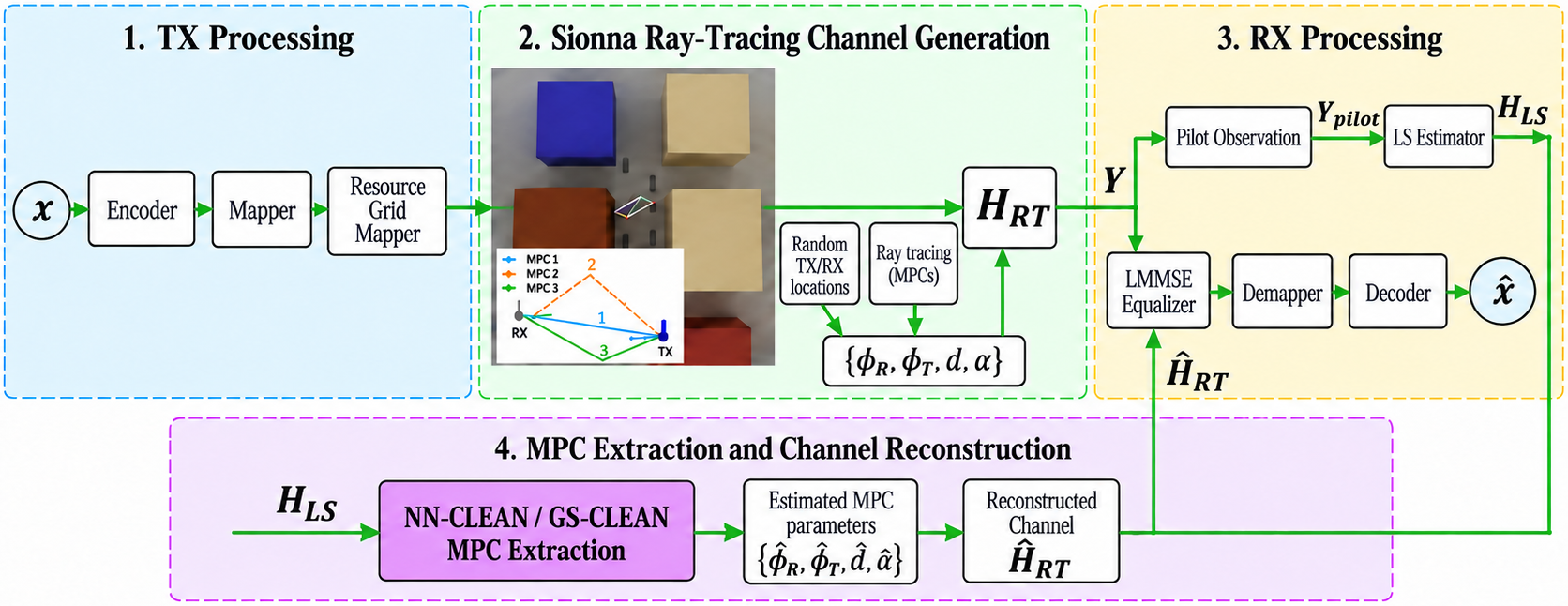}
    \caption{Sionna ray-tracing and communication-link evaluation framework.}
    \label{fig:rt_comm_framework}
\end{figure*}

A separate system-level evaluation is conducted using the ray-tracing
capabilities of the NVIDIA Sionna RT package~\cite{sionna}. The
\textit{simple\_street\_canyon\_with\_cars} scene is used to evaluate the
proposed method under geometry-dependent propagation within an end-to-end
communication link. The carrier frequency, antenna configuration, and
parameter-search ranges remain consistent with
Table~\ref{tab:system_configuration}.

Separate LOS and NLOS test sets are generated in the same ray-tracing
scene. The Tx and Rx are randomly positioned at valid roadway locations,
and placements intersecting scene objects are discarded. Realizations
with an unobstructed direct path are assigned to the LOS set; those
without a direct path are assigned to the NLOS set. Sionna RT provides
the AoA, AoD, propagation delay $\tau$, and complex coefficient of each
resolved MPC, with propagation distance obtained as $d=c\tau$. After
removing components outside the adopted estimation ranges and redundant
paths, the main LOS and NLOS evaluations retain
$N_{\mathrm{MPC}}=2$ paths per realization. A separate three-MPC LOS
test set is generated for the ablation study in
Section~\ref{sec:ablation_setup}. The trained NN-CLEAN model and GS-CLEAN
search grid remain fixed across these evaluations.

Each operating point in the LOS and NLOS sweeps is evaluated over
$10$ independent Monte Carlo runs, with $1{,}000$ test samples per
propagation condition in each run. The methods use the same channel
realizations within each run, and the reported path accuracy,
channel-reconstruction MSE, and BER are averaged over the $10$ runs.

The orthogonal frequency-division multiplexing (OFDM) resource grid
contains $N_s=14$ OFDM symbols and $N_f=512$ subcarriers, with a
cyclic-prefix length of $64$ samples. OFDM symbol~7 is allocated to
orthogonal MIMO pilots. The information bits are encoded using a
rate-$1/2$ 5G low-density parity-check (LDPC) code and mapped to
quadrature phase-shift keying (QPSK) symbols. The communication evaluation
covers $E_b/N_0\in[-3,10]$~dB, with the AWGN variance determined from
the modulation order, coding rate, and cyclic-prefix overhead.

For each test sample, Sionna RT generates the noiseless frequency-domain
channel
$\boldsymbol{H}_{\mathrm{RT}}
\in\mathbb{C}^{N_R\times N_T\times N_f}$.
Let
$\boldsymbol{X}\in\mathbb{C}^{N_T\times N_s\times N_f}$
and
$\boldsymbol{Y}\in\mathbb{C}^{N_R\times N_s\times N_f}$
denote the transmitted and received resource grids, respectively.
Their $k$th subcarrier slices are denoted by
$\boldsymbol{X}_k\in\mathbb{C}^{N_T\times N_s}$ and
$\boldsymbol{Y}_k\in\mathbb{C}^{N_R\times N_s}$.
Under the quasi-static channel assumption, the received signal on
each subcarrier is

\begin{equation}
\label{eq:rt_received_signal}
\boldsymbol{Y}_k
=
\boldsymbol{H}_{\mathrm{RT}}(f_k)\boldsymbol{X}_k
+
\boldsymbol{W}_k,
\qquad k=1,\ldots,N_f,
\end{equation}
where
$\boldsymbol{H}_{\mathrm{RT}}(f_k)
\in\mathbb{C}^{N_R\times N_T}$
is the channel matrix at frequency $f_k$, and
$\boldsymbol{W}_k\in\mathbb{C}^{N_R\times N_s}$ represents AWGN.
% The channel remains constant over the $N_s$ OFDM symbols.

The complete ray-tracing, channel-estimation, reconstruction, and
communication-receiver chain is illustrated in
Fig.~\ref{fig:rt_comm_framework}. The pilot-bearing entries of
$\boldsymbol{Y}$, denoted by $\boldsymbol{Y}_{\mathrm{pilot}}$, are
processed by the Least Squares (LS) channel estimator. Nearest-neighbor interpolation
produces the full-grid estimate
$\hat{\boldsymbol{H}}_{\mathrm{LS}}$, which serves as the common input to
NN-CLEAN and GS-CLEAN. The parameter--gain pairs estimated by each method
are then used to reconstruct $\hat{\boldsymbol{H}}$ according to the
channel expansion in \eqref{eq:channel_model}, with the additive-noise term
omitted.

The reconstructed channel $\hat{\boldsymbol{H}}$ and received resource grid
$\boldsymbol{Y}$ are supplied to the downstream receiver. The linear minimum mean-square error (LMMSE) equalizer produces
estimates of the transmitted QPSK symbols, which are subsequently
processed by the QPSK demapper and 5G LDPC decoder to obtain the
estimated information bits $\hat{\boldsymbol{x}}$.

The synthetic evaluation is parameterized by
$\mathrm{SNR}_{\mathrm{syn}}$, whereas the communication-link evaluation
uses $E_b/N_0$, with $E_b$ denoting the energy per information bit.
For the adopted OFDM configuration, pilot normalization, and noise
scaling, the measured pilot-level channel-observation SNR is
approximately $2$~dB below $E_b/N_0$ in both LOS and NLOS conditions.
Thus, $E_b/N_0=7$~dB corresponds to approximately $5$~dB at the pilots.
This correspondence provides a reference between the noise levels;
the LS channel estimation and interpolation used to form the CLEAN input
produce a different error structure from the directly perturbed
synthetic channel observations.

RT path accuracy is evaluated separately for the LOS and NLOS datasets
using the same extraction-order association and $\pm2$-grid tolerance adopted
for the synthetic evaluation, with \eqref{eq:path_accuracy} used to
compute the reported accuracy. All estimated MPCs are retained for
channel reconstruction, irrespective of whether they satisfy the
path-accuracy tolerance. The reconstructed channel is compared with the
noiseless RT reference using

\begin{equation}
\label{eq:channel_mse}
\mathrm{MSE}_{H}
=
\frac{1}{N_fN_TN_R}
\sum_{k=1}^{N_f}
\left\|
\boldsymbol{H}_{\mathrm{RT}}(f_k)
-
\hat{\boldsymbol{H}}(f_k)
\right\|_F^2.
\end{equation}

Communication performance is characterized by the BER obtained by comparing
the decoded information bits $\hat{\boldsymbol{x}}$ with the transmitted
information bits $\boldsymbol{x}$ at each $E_b/N_0$ operating point. LS
channel estimation with nearest-neighbor interpolation and perfect channel
state information (CSI) are included as the conventional and idealized
receiver references, respectively.

\subsubsection{Ablation Study}
\label{sec:ablation_setup}

The full NN-CLEAN model trained according to
Algorithm~\ref{alg:model-training} is compared with four ablation
variants. In \emph{w/o-GLS}, the targets in \eqref{eq:gls} are replaced
by one-hot targets at the same ground-truth class indices, while the CCE
in \eqref{eq:cce_definition} is retained. In \emph{w/o-gating}, the
distance head receives the shared backbone features through a branch
parallel to the AoA and AoD heads, without angular conditioning. The
\emph{uniform-weight} variant changes the distance CCE coefficient in
\eqref{eq:step_loss} from $3$ to $1$, assigning equal weight to all three
parameter losses. The \emph{direct-Phase3} variant uses the Phase~3
residual updates and loss in \eqref{eq:phase_loss} from the first
training step, omitting Phases~1 and~2.

Except for the stated modification, each variant retains the full
model's architecture and training settings described in
Section~\ref{sec:training_setup}, including the same total number of
optimizer updates. All models are trained separately and evaluated
on the same three-MPC LOS RT test samples using the receiver
configuration described above. Path accuracy, channel-reconstruction
MSE, and BER are compared at $E_b/N_0=5$~dB.

\section{Results and Discussion}
\label{sec:results}

\subsection{Baseline Multipath Estimation Performance}

\begin{figure}[htbp]
  \centering
  \includegraphics[width=\columnwidth]
  {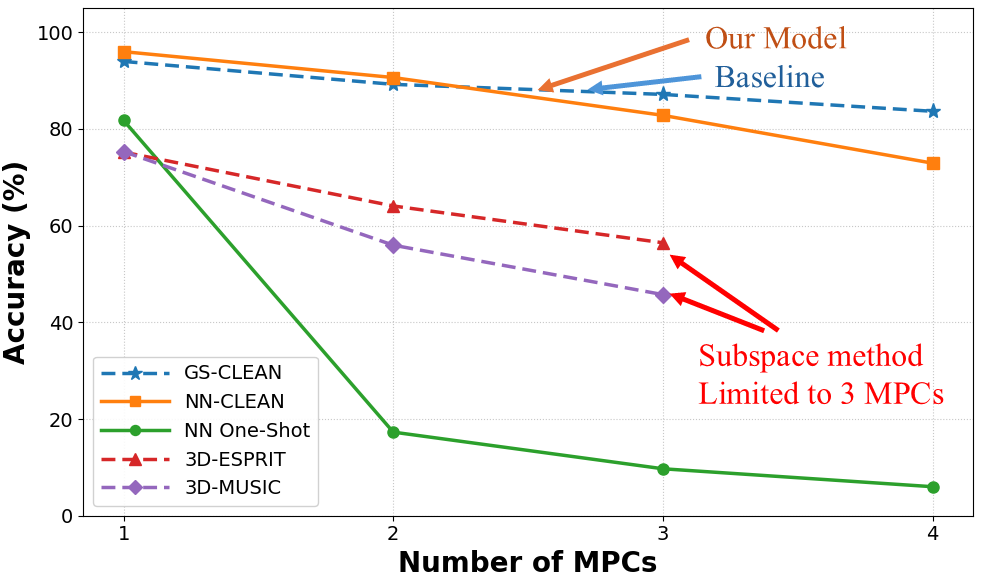}
  \caption{Off-grid MPC estimation accuracy versus path count under the
  baseline synthetic configuration at
  $\mathrm{SNR}_{\mathrm{syn}}=5$~dB.}
  \label{fig:accuracy_mpc}
\end{figure}

Under the off-grid synthetic evaluation,
Fig.~\ref{fig:accuracy_mpc} compares NN-CLEAN with 
GS-CLEAN \cite{seun_clean}, 3D-MUSIC \cite{3d_music}, 3D-ESPRIT \cite{3d_esprit}, and the one-shot NN as the path count increases. For
$N_{\mathrm{MPC}}\in\{1,2\}$, NN-CLEAN achieves the highest path accuracy,
slightly exceeding GS-CLEAN while maintaining a clear advantage over the
subspace and one-shot baselines. In these sparse cases, the learned
estimator identifies the relevant region of the three-dimensional
AoA--AoD--distance space without explicitly evaluating every candidate at
each CLEAN iteration.

For $N_{\mathrm{MPC}}\in\{3,4\}$, the accuracy of both CLEAN-based
methods decreases, with GS-CLEAN retaining higher accuracy.
Because the reference parameters are off-grid, reconstructing a path
with a discrete dictionary atom generally leaves structured residual
energy after subtraction. These errors affect subsequent extraction
steps and make weaker MPCs more difficult to recover. At a fixed
aggregate $\mathrm{SNR}_{\mathrm{syn}}$, increasing the path count
also introduces greater inter-path interference and can reduce the
individual SNR of weaker components. NN-CLEAN additionally depends
on learned parameter decisions, and an inaccurate early estimate can
distort the residual supplied to later iterations. Its larger accuracy
reduction is consistent with greater sensitivity to these increasingly
difficult residual observations.

The one-shot NN degrades more rapidly, even with ideal cross-MPC
association. It predicts every MPC from the original superposition,
whereas NN-CLEAN updates the observation after each estimated path
is removed. Successive cancellation reduces the contribution of
previously extracted components before later predictions. The
comparison therefore supports retaining the CLEAN recursion under
the evaluated baseline configuration.

\subsection{Parameter Precision and Reconstruction Fidelity}

\begin{figure*}[t!]
  \centering
  \begin{subfigure}[b]{0.32\linewidth}
    \centering
    \includegraphics[width=\linewidth]{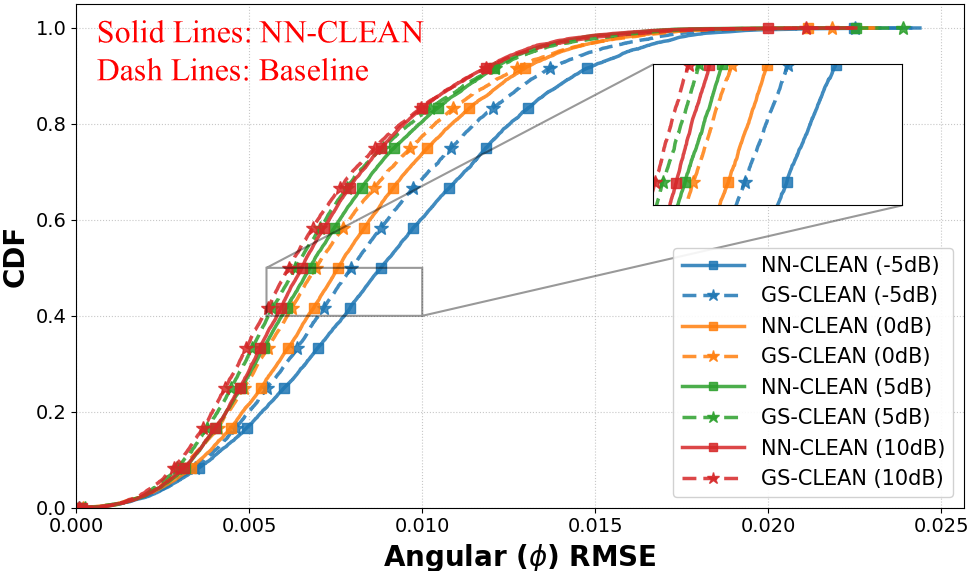}
    \caption{Angular MCD.}
    \label{fig:angle_mcd}
  \end{subfigure}
  \hfill
  \begin{subfigure}[b]{0.32\linewidth}
    \centering
    \includegraphics[width=\linewidth]{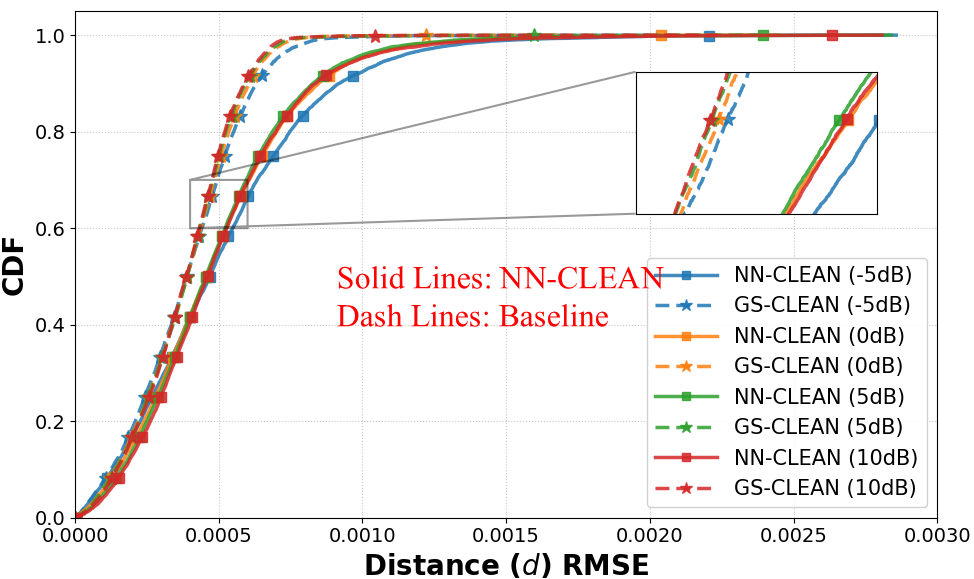}
    \caption{Distance MCD.}
    \label{fig:distance_mcd}
  \end{subfigure}
  \hfill
  \begin{subfigure}[b]{0.32\linewidth}
    \centering
    \includegraphics[width=\linewidth]
    {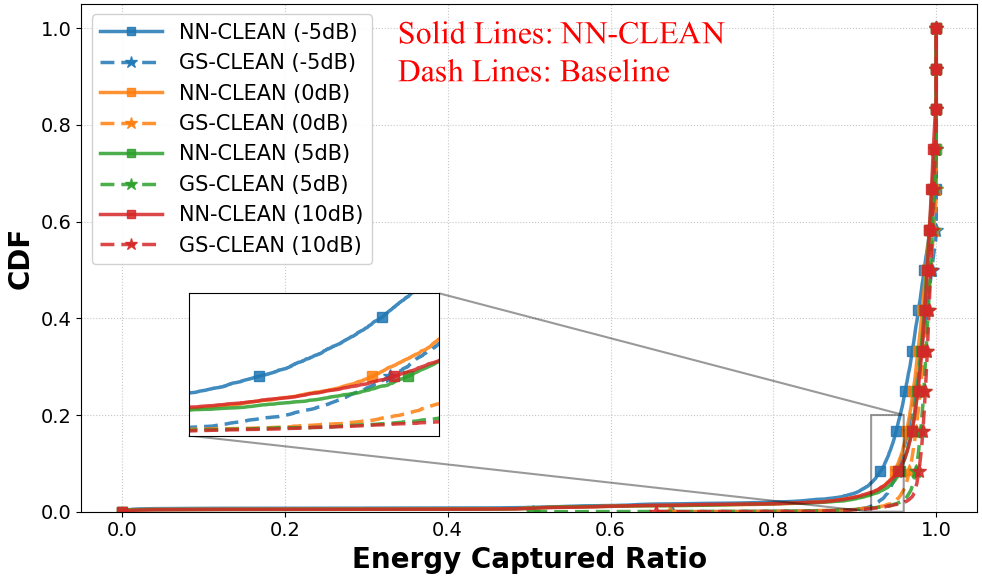}
    \caption{Energy-captured ratio.}
    \label{fig:energy_ratio_result}
  \end{subfigure}
  \caption{CDFs of angular MCD, distance MCD, and energy-captured ratio for
  $N_{\mathrm{MPC}}=2$ over the synthetic SNR conditions.}
  \label{fig:parameter_precision}
\end{figure*}

Path accuracy determines whether an MPC satisfies the prescribed detection
tolerance but does not characterize the residual error of a successfully
associated path. Fig.~\ref{fig:parameter_precision} therefore presents the
empirical cumulative distribution functions (CDFs) of angular MCD
(Fig.~\ref{fig:angle_mcd}), distance MCD
(Fig.~\ref{fig:distance_mcd}), and energy-captured ratio
(Fig.~\ref{fig:energy_ratio_result}) for the two-path case. The MCD
distributions are conditional on successful path association and therefore
measure estimation precision rather than unconditional path-recovery
performance.

As $\mathrm{SNR}_{\mathrm{syn}}$ decreases, the angular- and distance-error
distributions broaden, while the energy-captured ratio deviates further from
its ideal value. GS-CLEAN retains the highest conditional precision because
it evaluates the complete candidate grid explicitly. Nevertheless, the
NN-CLEAN distributions remain close to the GS-CLEAN reference across the
evaluated SNR range, indicating that the additional localization error is
limited once NN-CLEAN identifies the correct MPC neighborhood.

Both methods remain subject to the same off-grid quantization floor because
the continuously generated paths are reconstructed using discrete
dictionary atoms. Angular mismatch perturbs the spatial array response,
whereas distance mismatch produces a coherent phase error across the
observed bandwidth. The latter persists over all subcarriers and can
propagate through successive residual updates. Although NN-CLEAN introduces
an additional classification error, the closely aligned MCD and
energy-captured distributions show that it preserves most of the dominant
geometric and energetic channel structure.

\subsection{Runtime and Memory Efficiency}
\label{sec:runtime_results}

\begin{figure}[htbp]
  \centering
  \includegraphics[width=\columnwidth]{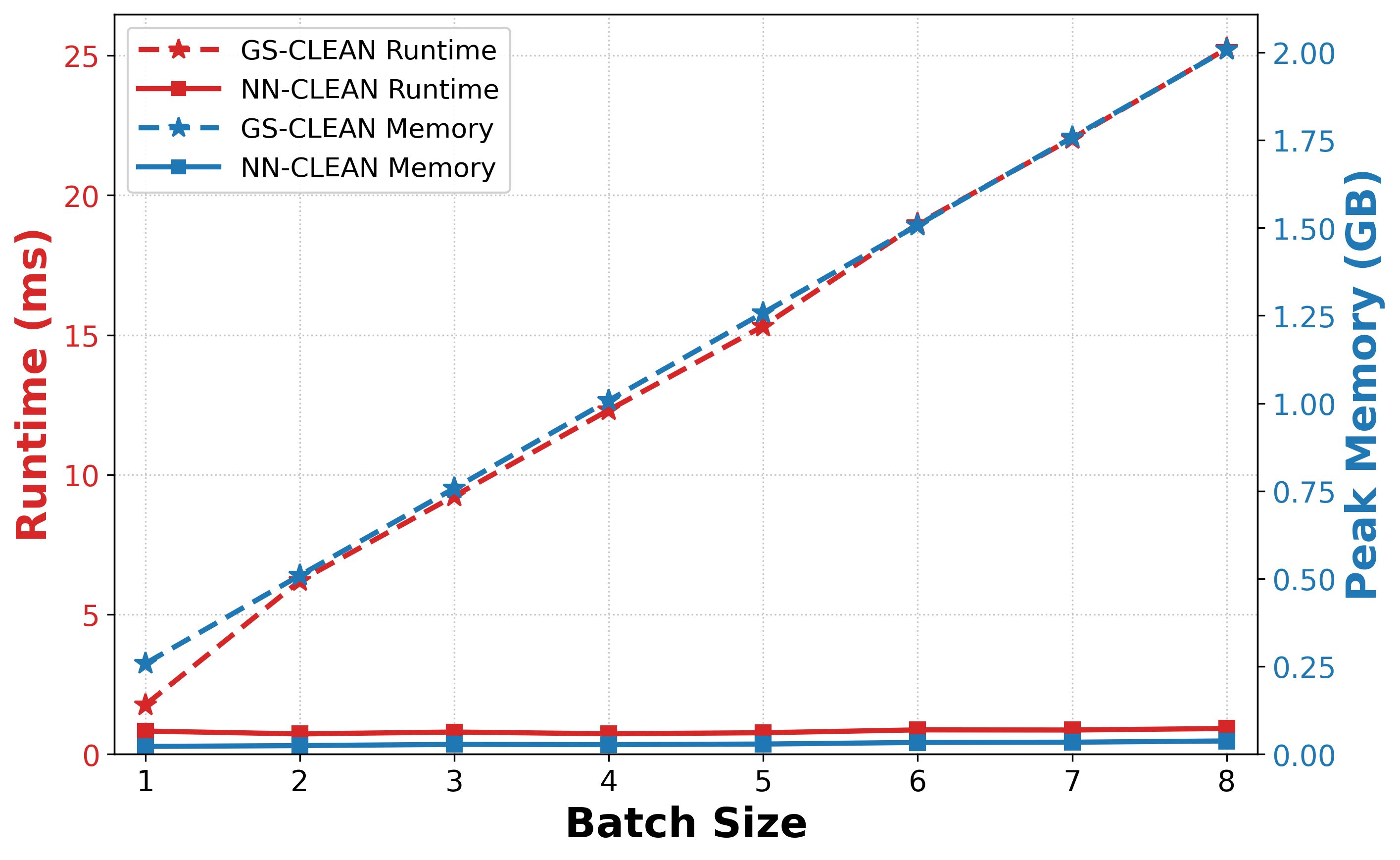}
  \caption{Runtime and Memory Comparison of NN-CLEAN and GS-CLEAN for one-MPC}
  \label{fig:runtime_memory}
\end{figure}

Fig.~\ref{fig:runtime_memory} compares the runtime and peak GPU memory
of the one-MPC parameter-search stage as the processing batch size,
$N_{\mathrm{batch}}$, increases from $1$ to $8$. At
$N_{\mathrm{batch}}=1$, NN-CLEAN requires $0.82$~ms, compared to
$1.76$~ms for GS-CLEAN. GS-CLEAN runtime and memory increase
approximately linearly over the batch sweep, with both quantities
roughly doubling from $N_{\mathrm{batch}}=4$ to $8$. NN-CLEAN
runtime remains below $1$~ms throughout the sweep, while its memory
requirement grows much more slowly. At $N_{\mathrm{batch}}=8$,
NN-CLEAN takes $0.92$~ms, compared to $25.24$~ms for GS-CLEAN,
corresponding to an approximately $28\times$ speedup. Peak GPU
memory is $0.04$~GB for NN-CLEAN and $2.16$~GB for GS-CLEAN,
a reduction by a factor of approximately $53$. The computational
advantage therefore becomes substantially larger as more channels
are processed within the same estimation call.

\begin{table}[t]
\centering
\caption{Runtime and Memory Comparison of NN-CLEAN and GS-CLEAN with three-MPC}
\label{tab:runtime_memory}
\footnotesize
\setlength{\tabcolsep}{3pt}
\renewcommand{\arraystretch}{1.10}
\begin{tabular}{clccc}
\hline
$N_{\mathrm{batch}}$ & \textbf{Method} &
\textbf{\shortstack{One-MPC\\search (ms)}} &
\textbf{\shortstack{Three-MPC\\end-to-end\\runtime (ms)}} &
\textbf{\shortstack{Three-MPC\\peak memory\\(GB)}} \\
\hline
1 & GS-CLEAN & 1.76 & 5.40 & 0.28 \\
1 & NN-CLEAN & 0.82 & 1.91 & 0.02 \\
\hline
4 & GS-CLEAN & 12.31 & 36.68 & 1.08 \\
4 & NN-CLEAN & 0.73 & 2.14 & 0.03 \\
\hline
8 & GS-CLEAN & 25.24 & 75.16 & 2.16 \\
8 & NN-CLEAN & 0.92 & 2.80 & 0.04 \\
\hline
\end{tabular}
\end{table}

Table~\ref{tab:runtime_memory} compares the one-MPC search times
with the end-to-end runtime and peak memory of complete three-MPC
extraction. GS-CLEAN takes approximately $3\times$ its one-MPC
search time, consistent with dictionary evaluation dominating
the workload and being repeated at each extraction step.
NN-CLEAN requires $1.91$~ms at $N_{\mathrm{batch}}=1$,
approximately $2.3\times$ its one-MPC search time. This lower
ratio is consistent with fixed call and synchronization overhead
forming a larger fraction of the short single-path measurement
and being amortized within the three-path estimation call.
The ratio approaches $3\times$ as the batch workload increases.
At $N_{\mathrm{batch}}=8$, complete NN-CLEAN extraction achieves
an approximately $27\times$ speedup and a $52$-fold reduction
in peak GPU memory compared to GS-CLEAN. For both methods,
peak memory remains close to the corresponding one-MPC requirement
because the processing workspace is reused between successive
extraction steps.

\subsection{Impact of Observation Dimensions and Search-Grid Resolution}
\label{sec:dimension_results}
\begin{figure*}[t]
  \centering

  \begin{subfigure}[b]{0.32\linewidth}
    \centering
    \includegraphics[width=\linewidth]
    {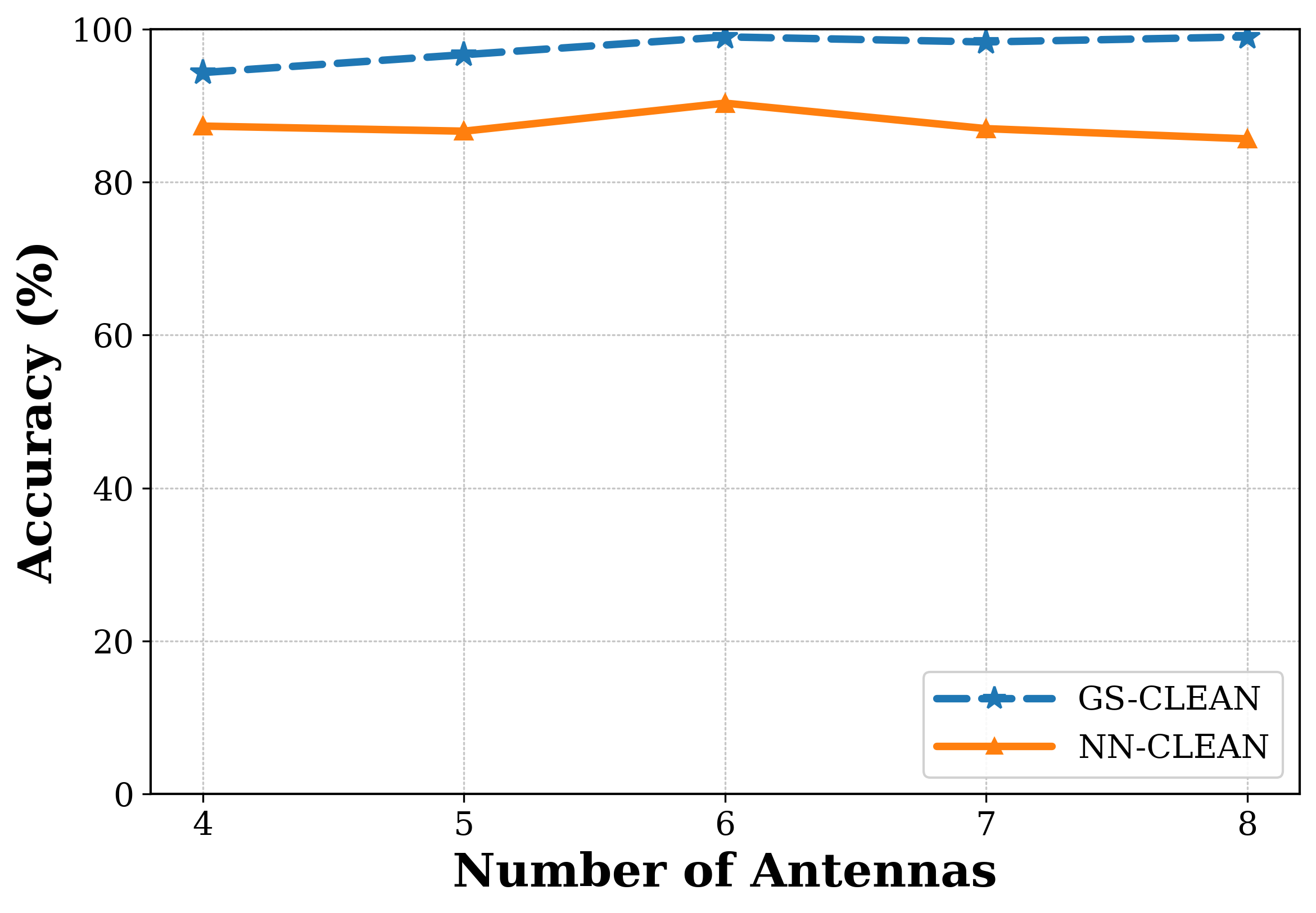}
    \caption{Antenna-number sweep.}
    \label{fig:antenna_sweep}
  \end{subfigure}
  \hfill
  \begin{subfigure}[b]{0.32\linewidth}
    \centering
    \includegraphics[width=\linewidth]
    {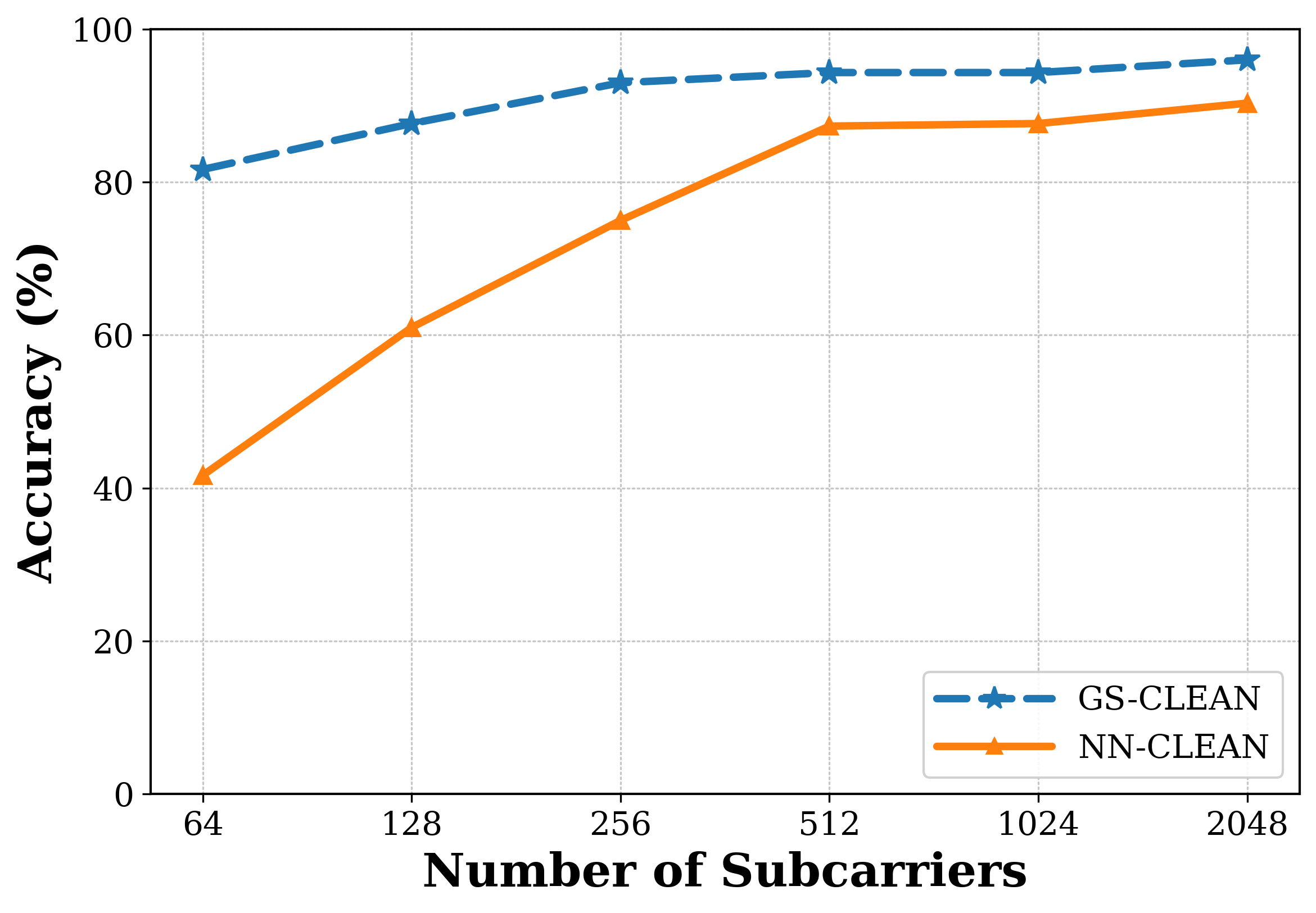}
    \caption{Subcarrier-number sweep.}
    \label{fig:subcarrier_sweep}
  \end{subfigure}
  \hfill
  \begin{subfigure}[b]{0.32\linewidth}
    \centering
    \includegraphics[width=\linewidth]
    {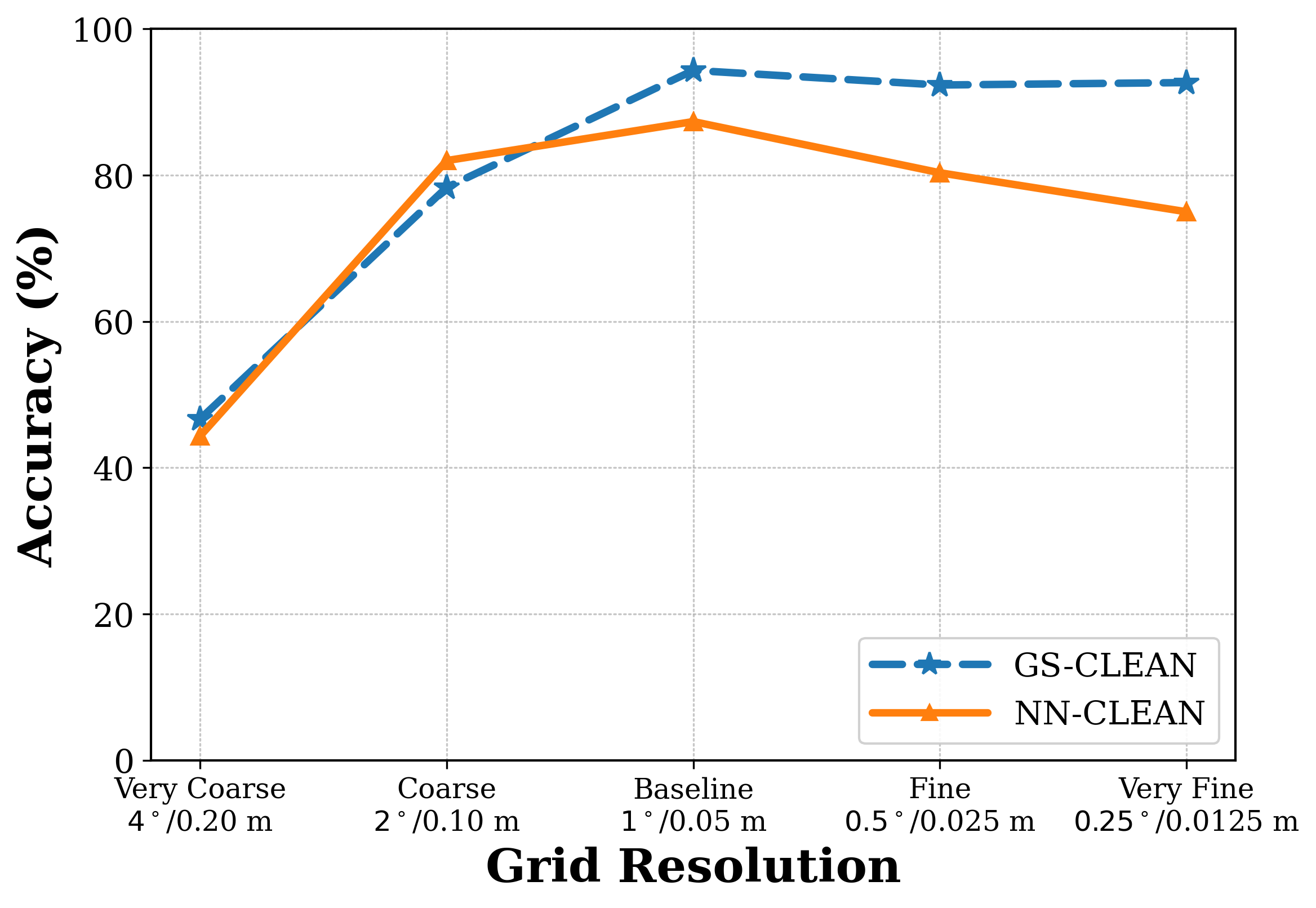}
    \caption{Grid-resolution sweep.}
    \label{fig:resolution_sweep}
  \end{subfigure}

  \caption{Accuracy under variations in  observation
  dimension and search grid resolution
  for $N_{\mathrm{MPC}}=3$ at $\mathrm{SNR}_{\mathrm{syn}}=5$~dB.}
  \label{fig:dimension_sweep}
\end{figure*}

Fig.~\ref{fig:dimension_sweep} compares the effects of observation
dimensions and search-grid resolution on three-MPC off-grid estimation
at $\mathrm{SNR}_{\mathrm{syn}}=5$~dB. In
Fig.~\ref{fig:dimension_sweep}(a), increasing $N_T=N_R$ from $4$ to
$8$ gives GS-CLEAN a modest accuracy improvement, while NN-CLEAN
shows no consistent increase. The number of frequency observations
has a stronger effect, as shown in Fig.~\ref{fig:dimension_sweep}(b).
Both methods improve with $N_f$, and NN-CLEAN obtains most of its
gain below the baseline $N_f=512$. Additional frequency observations
help constrain the distance-dependent phase progression used in
parameter estimation and cancellation. Because the bandwidth is
fixed, increasing $N_f$ provides denser sampling and supports noise
averaging without changing the nominal delay resolution. The gains
become smaller beyond $N_f=512$, which captures most of the
improvement observed in this sweep.

Fig.~\ref{fig:dimension_sweep}(c) compares search grids using the same
physical accuracy tolerances. NN-CLEAN slightly exceeds GS-CLEAN
on the coarse grid, while both methods achieve their highest measured
accuracy at the baseline resolution. Finer grids provide no consistent
improvement for GS-CLEAN and reduce NN-CLEAN accuracy. For NN-CLEAN,
the classification outputs and their directly connected layers expand
with grid refinement, but the shared backbone, hidden feature dimensions,
and optimization-step count remain fixed. The model must therefore
distinguish more closely spaced classes using the same feature
representation. In addition, the fixed GLS width in bins corresponds
to a narrower physical smoothing interval on finer grids. These
training and architecture choices may limit the benefit of reducing
the grid spacing.

\subsection{Ray-Tracing Generalization and Communication Performance}
The Sionna RT experiments provide a geometry-based validation of NN-CLEAN
and evaluate how parameter-estimation errors affect channel reconstruction
and downstream communication performance. Unlike the independently sampled
synthetic parameters, the RT paths are determined jointly by the scene
geometry and propagation interactions. The following results therefore
evaluate both the transfer of the learned search to a different channel
distribution and the effect of the resulting estimation errors on the
decoded communication link.

\subsubsection{Ray-Tracing MPC Estimation Accuracy}

\begin{figure}[htbp]
  \centering
  \includegraphics[width=\columnwidth]{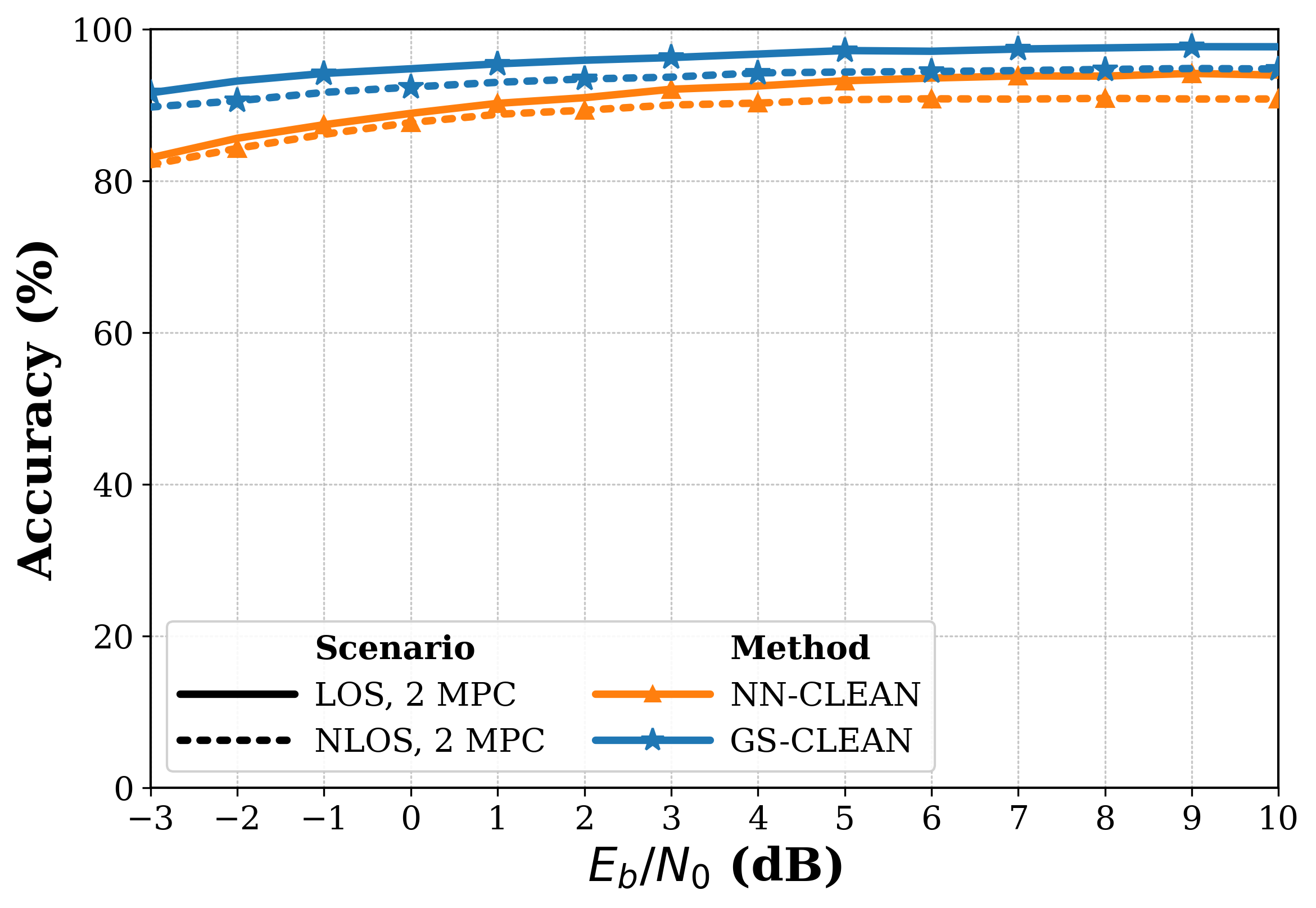}
  \caption{Accuracy versus $E_b/N_0$ for two-MPC LOS and
  NLOS RT channels.}
  \label{fig:rt_accuracy}
\end{figure}

Fig.~\ref{fig:rt_accuracy} compares path accuracy for the two-MPC
LOS and NLOS test sets. Accuracy generally increases with $E_b/N_0$
and levels off toward the upper end of the sweep. GS-CLEAN maintains
higher accuracy in both propagation conditions, while NN-CLEAN
saturates earlier, particularly for NLOS channels. Both estimators
achieve higher accuracy in LOS compared to NLOS under the same
retained path count, search grid, and evaluation tolerance.

The RT result differs from the synthetic two-MPC baseline, where
NN-CLEAN slightly exceeds GS-CLEAN. RT paths have geometry-dependent
parameter relationships, and their channel observations are obtained
through pilot-based LS channel estimation and interpolation. Applying the
trained NN-CLEAN model without retraining therefore exposes it to
changes in both path statistics and input-error structure.
GS-CLEAN continues to evaluate the physical dictionary directly.
The observed accuracy gap captures the combined change in channel
distribution and observation pipeline; the pilot-level SNR
correspondence does not establish identical CLEAN input conditions.

\subsubsection{Ray-Tracing Channel Reconstruction}

\begin{figure}[htbp]
  \centering
  \includegraphics[width=\columnwidth]{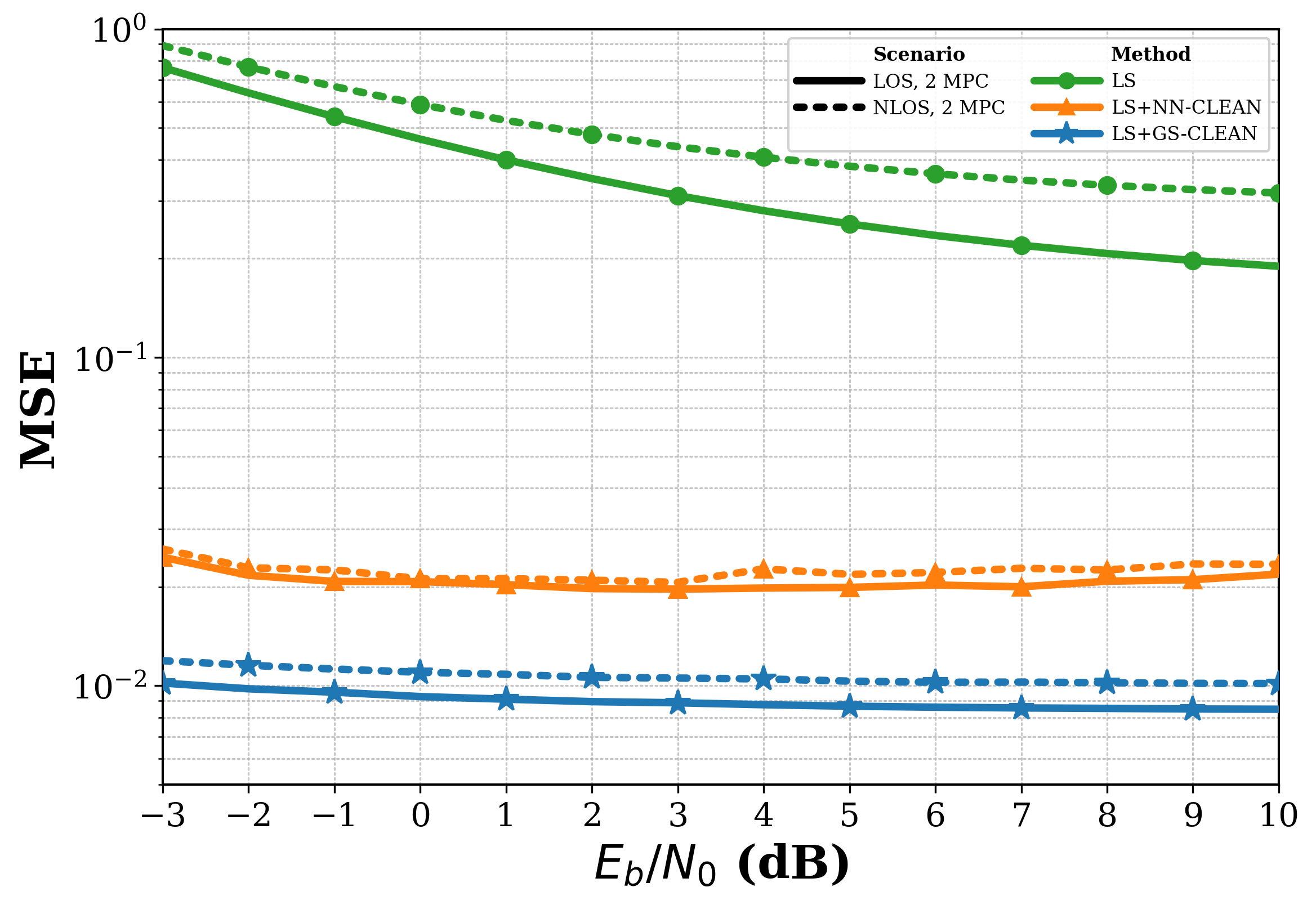}
  \caption{Channel-reconstruction MSE versus $E_b/N_0$ for two-MPC LOS and
  NLOS RT channels.}
  \label{fig:rt_mse}
\end{figure}

Fig.~\ref{fig:rt_mse} compares channel-reconstruction MSE against the
noiseless two-MPC RT reference. Both CLEAN-based methods achieve
substantially lower MSE compared to LS channel estimation with nearest-neighbor
interpolation. The parametric fit constrains the reconstructed channel
to the spatial--frequency responses of the estimated paths, reducing
the error of the interpolated observation. GS-CLEAN achieves the lowest
MSE in both propagation conditions. Over the tested $E_b/N_0$ range,
the maximum absolute difference between NN-CLEAN and GS-CLEAN remains
below $1.4\times10^{-2}$ for both LOS and NLOS channels.

All estimated MPCs are included in reconstruction, weighted by their
complex gains. An estimate outside the path-accuracy tolerance can
still approximate part of the reference response, and its effect
on MSE depends on its power and the resulting spatial--frequency
mismatch. At higher $E_b/N_0$, both reconstruction curves vary only
weakly as pilot noise decreases. This behavior is consistent with
persistent errors from grid discretization and the interpolated LS
input. These errors affect the fitted parameters and gains and can
propagate through successive cancellation, limiting further
improvement under the fixed estimation pipeline.

\subsubsection{End-to-End Communication Performance}

\begin{figure}[htbp]
  \centering
  \includegraphics[width=\columnwidth]{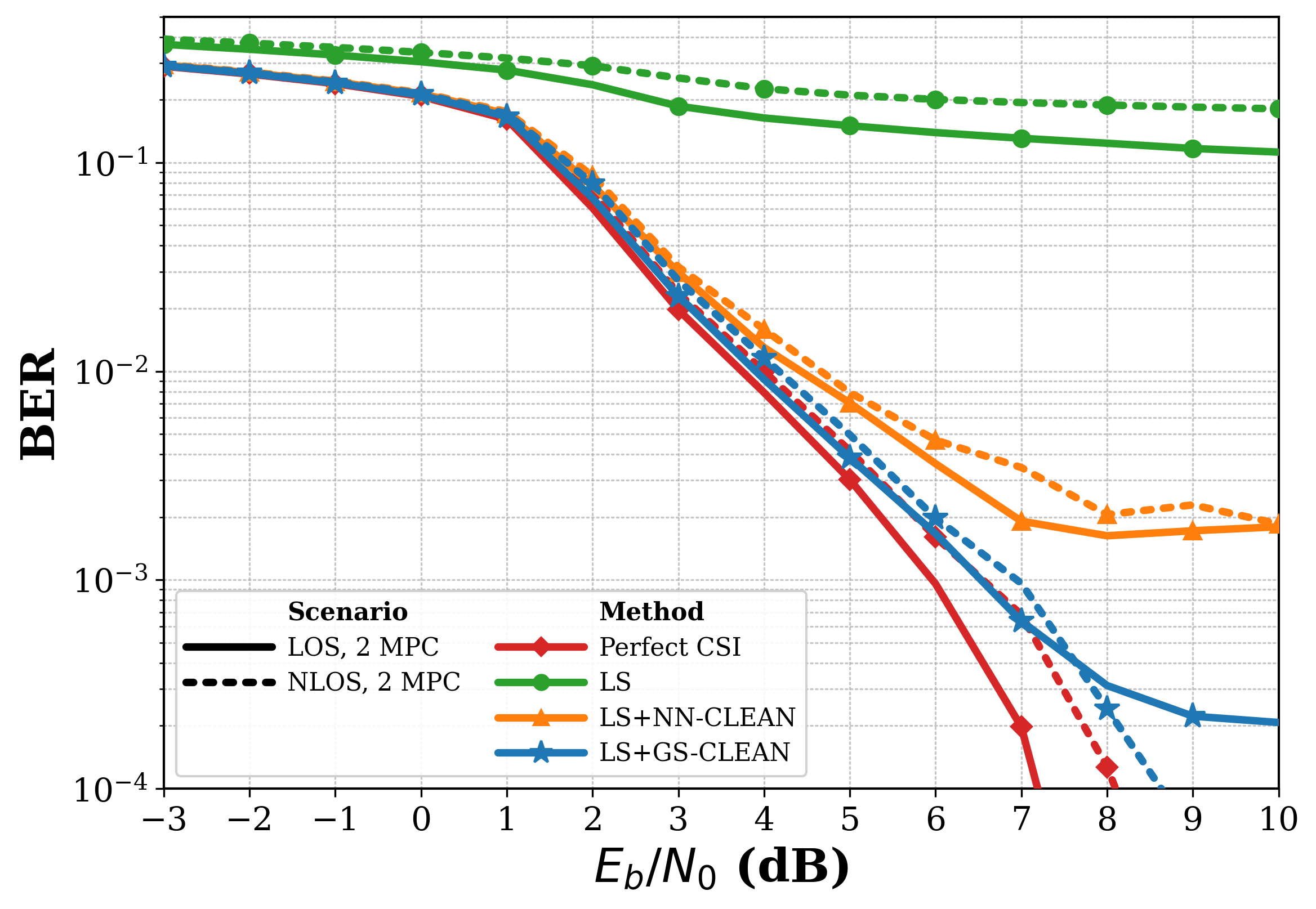}
  \caption{BER versus $E_b/N_0$ for two-MPC LOS and
  NLOS RT channels.}
  \label{fig:rt_ber}
\end{figure}

Fig.~\ref{fig:rt_ber} compares BER for the four receiver cases in
two-MPC LOS and NLOS channels. Perfect CSI provides the lowest BER,
followed by GS-CLEAN, NN-CLEAN, and LS channel estimation with nearest-neighbor
interpolation. Both CLEAN-based reconstructions substantially improve
BER compared to LS in both propagation conditions. LOS generally
provides lower BER compared to NLOS across the four receivers,
apart from a small GS-CLEAN reversal at the upper end of the sweep.
The LOS advantage also appears with perfect CSI, showing that the
LOS--NLOS difference reflects the communication characteristics of
the channels as well as channel-estimation error.

Over the entire tested $E_b/N_0$ range, the maximum absolute BER
difference between NN-CLEAN and GS-CLEAN is approximately $1.05\%$
for LOS and $0.69\%$ for NLOS. At $E_b/N_0=5$~dB, the absolute
NN-CLEAN--perfect-CSI gap is approximately $0.4\%$ in both conditions.
NN-CLEAN BER levels off toward the upper end of the sweep, consistent
with its residual channel-reconstruction error. Its reconstruction
nevertheless retains a substantial BER advantage over interpolated
LS, while the gap to GS-CLEAN quantifies the link-performance penalty
associated with the computational savings.

\subsubsection{Ablation Study}
\label{sec:ablation_results}

Table~\ref{tab:rt_ablation} compares the full model and four variants
on three-MPC LOS RT channels at $E_b/N_0=5$~dB. The
\emph{w/o-GLS} variant shows the largest deterioration across all
three metrics. This result supports distributing supervision across
neighboring parameter bins, whose corresponding channel responses
are correlated. Removing angular conditioning in
\emph{w/o-gating} produces a smaller loss in path accuracy, together
with higher reconstruction MSE and BER.

\begin{table}[t]
\centering
\caption{NN-CLEAN Ablation Results for Three-MPC LOS RT Channels
at $E_b/N_0=5$~dB}
\label{tab:rt_ablation}
\footnotesize
\renewcommand{\arraystretch}{1.10}
\begin{tabular}{lccc}
\hline
\textbf{Variant} & \textbf{Accuracy (\%)} &
\textbf{MSE} & \textbf{BER (\%)} \\
\hline
\textbf{Full NN-CLEAN} & \textbf{75.13} &
\textbf{\boldmath$2.51\times10^{-2}$} & \textbf{0.79} \\
w/o-GLS & 71.24 & $3.48\times10^{-2}$ & 1.40 \\
w/o-gating & 74.53 & $2.61\times10^{-2}$ & 0.89 \\
uniform-weight & 75.12 & $3.08\times10^{-2}$ & 1.11 \\
direct-Phase3 & 74.17 & $2.67\times10^{-2}$ & 1.01 \\
\hline
\end{tabular}
\end{table}

The \emph{uniform-weight} variant retains nearly the same path
accuracy as the full model but has higher MSE and BER. The larger
distance-loss weight therefore benefits reconstruction and link
performance even when aggregate path accuracy changes little.
This outcome is consistent with distance errors introducing
frequency-dependent phase mismatch in the reconstructed paths and
subsequent residuals. The \emph{direct-Phase3} variant also degrades
all three metrics, supporting the gradual introduction of sequential
extraction before training with prediction-dependent residuals.
This comparison assesses the complete curriculum and does not
isolate the individual contributions of Phases~1 and~2.

\section{Conclusion}

This paper presented NN-CLEAN, which replaces the exhaustive
parameter search within CLEAN with a multi-head neural classifier
while retaining LS gain estimation, parametric reconstruction,
and residual subtraction. Reusing the same single-MPC network
permits different extraction depths without changing its output
structure. Off-grid synthetic experiments show that NN-CLEAN
is most competitive with GS-CLEAN in sparse channels, while
GS-CLEAN retains higher accuracy as the path count increases.
The observation-dimension and grid-resolution sweeps also show
that larger antenna configurations and finer output grids do
not uniformly improve the learned estimator.

For complete three-MPC extraction at a batch size of eight,
NN-CLEAN achieves an approximately $27\times$ speedup and a
$52$-fold reduction in peak GPU memory compared to GS-CLEAN.
The two-MPC RT evaluation shows a reconstruction and BER penalty
relative to GS-CLEAN in both LOS and NLOS, while both CLEAN-based
methods substantially improve performance over interpolated LS.
At $E_b/N_0=5$~dB, the absolute BER gap between NN-CLEAN and
perfect CSI is approximately $0.4\%$ in both conditions.
Three-MPC LOS ablations support label smoothing, angular
conditioning, distance-loss weighting, and staged training.
The results quantify the tradeoff between batched computational
efficiency and estimation performance. Adaptive stopping for
unknown path counts remains a direction for future work.

\section{ACKNOWLEDGEMENTS}
This work was supported in part by CoCoSys, one of seven
centers in JUMP 2.0, a Semiconductor Research Corporation
(SRC) program sponsored by DARPA.

\bibliographystyle{IEEEtran}
\bibliography{refs}

\end{document}